\documentclass[11pt,twoside]{article}

\usepackage{amsthm}
\usepackage{amsmath}
\usepackage{authblk}
\usepackage{float}
\usepackage{bbm}
\usepackage{paralist}
\usepackage{hyperref}
\usepackage[pdftex]{graphicx}
\usepackage[left=1.25in, top=1.25in, right=1.25in, bottom=1.25in]{geometry}
\usepackage{algpseudocode}
\usepackage{tikz}
\usepackage{subfig}
\usepackage{adjustbox}
\usepackage[toc,page]{appendix}

\usepackage{fancyhdr} 
\definecolor{darkgreen}{rgb}{0.0, 0.5, 0.0}

\theoremstyle{plain}
\newtheorem{thm}{Theorem}[section]
\theoremstyle{definition}
\newtheorem{defn}[thm]{Definition}

\begin{document}

\title{Enhancing AI-Based ECG Delineation with Deep Learning Denoising Techniques}
\date{\today}

\author[1]{Jeff Breeding-Allison}
\author[1]{Emil Walleser}
\affil[1]{Mars Petcare}

\maketitle

\begin{abstract}
Evaluating canine electrocardiograms (ECGs) is challenging due to noise that can obscure clinically relevant cardiac electrical activity. Common sources of interference include respiration, muscle activity, poor lead contact, and external electrical artifacts. Classical signal denoising techniques, such as filtering and wavelet-based methods, struggle to suppress diverse noise patterns while preserving morphological features critical for accurate ECG delineation. We propose an autoencoder-based neural network model and training strategy for ECG denoising as a preprocessing step for canine ECG analysis. The model is trained to reconstruct clean cardiac signals from noisy inputs, enabling effective noise reduction without degrading diagnostically important waveforms. Our approach demonstrates strong performance across both noisy and clean ECG recordings, indicating robustness to varying signal conditions and suitability for downstream delineation tasks.
\end{abstract}


\tableofcontents

%%%%%%%%%%%%%%%%%%%%%%%%%%%%%%%%%%%%%%%%%%
\section{Introduction}\label{intro}
%%%%%%%%%%%%%%%%%%%%%%%%%%%%%%%%%%%%%%%%%%

Electrocardiograms (ECGs) play a critical role in assessing cardiac health \cite{santilli2018electrocardiography}. However, ECG recordings from canine patients are often compromised by significant electrical noise. This interference, largely attributable to increased movement and restlessness compared with human patients, can obscure subtle cardiac electrical activity, thereby complicating accurate clinical interpretations.

Traditional signal-processing approaches, such as wavelet transforms and conventional filtering methods, are commonly employed for noise reduction in physiological signals. Many of these filters are available in the \verb"Neurokit2" Python package \cite{makowski2021neurokit}. Despite their utility, these methods have demonstrated limitations when applied to canine ECGs. They often fail to produce a sufficiently clean signal that concurrently preserves the critical morphological features of the heart's electrical activity necessary for diagnosis \cite{kramer2022ecgassess}, \cite{satija2018reviewsignalprocess}. This challenge has been the primary motivation for the current work, which focuses on developing a deep learning-based solution for robust canine ECG denoising.

The development of deep learning models for ECG analysis is an active area of research. Such models have been successfully applied to tasks including cardiomyopathy classification, arrhythmia detection, and other clinically relevant applications \cite{avula2023clinical}, \cite{dourson2023pulsenetdeeplearningecgsignal},  \cite{musa2022systematicreview}.

A significant hurdle in developing machine learning models for ECG denoising lies in constructing an appropriate training dataset consisting of noisy signals paired with clean reference recordings. This difficulty arises from several factors, including
\begin{itemize}
    \item the scarcity of truly noise-free data,
    \item the complex and variable nature of noise,
    \item the necessity for precise alignment between noisy and clean signal pairs, and
    \item the model's ability to generalize to unseen data.
\end{itemize}

In this study, we address these challenges by first curating a substantial dataset of $10,000$ canine ECG segments deemed sufficiently clean for use as reference signals. We then describe a methodology for generating realistic noisy counterparts, thereby constructing (noisy, clean) training examples. In addition, we explore the inclusion of (clean, clean) signal pairs during training to enhance model generalization and to establish a baseline for denoising performance. An efficient data-handling pipeline is implemented using PyTorch Datasets and Dataloaders \cite{paszke2019pytorch}.

Model performance is assessed by quantifying the similarity between the denoised outputs and the original signals, as well as by measuring improvements in classification accuracy for downstream ECG delineation tasks. These results are benchmarked against traditional signal-processing filters to demonstrate the advantages of the proposed deep learning approach.

%%%%%%%%%%%%%%%%%%%%%%%%%%%%%%%%%%%%%%%%%%
\section{Background}
%%%%%%%%%%%%%%%%%%%%%%%%%%%%%%%%%%%%%%%%%%

%%%%%%%%%%%%%%%%%%%%%
\subsection{The electrocardiogram (ECG)}
%%%%%%%%%%%%%%%%%%%%%

An \textbf{electrocardiogram} (ECG) is a non-invasive diagnostic tool that records the electrical activity of the heart by measuring changes in electrical potential on the body surface \cite{avula2023clinical}. It is widely used in both human and veterinary medicine for the assessment and diagnosis of numerous cardiovascular conditions.

The rhythmic contraction of the heart is governed by a specialized electrical conduction system. This system initiates and propagates electrical impulses in a coordinated manner, ensuring efficient blood flow through the heart and to the rest of the body. Alterations in cardiac structure or function are often reflected as changes in this electrical activity, which can be detected and analyzed via the ECG.

%%%%%%%%%%%%%%%%%%%%%
\subsection{ECG leads and recording}
%%%%%%%%%%%%%%%%%%%%%

To capture a comprehensive view of the heart's electrical activity, multiple ECG \textbf{leads}, representing distinct electrical viewpoints, are typically configured using electrodes placed at specific locations on the patient's body. Each lead measures the electrical potential difference between two electrodes, or between a single electrode and a reference, thereby providing a unique perspective on heart's electrical activity. Collectively, these leads offer complementary views that characterize the three-dimensional electrical events of the cardiac cycle. 

Beyond the directly acquired leads, additional ECG leads can be derived mathematically through linear combinations of the measured signals, such as by computing the difference between two leads. These derived leads can provide complementary diagnostic information and may reveal cardiac abnormalities that are less evident in the directly recorded waveforms. Standard ECG systems typically present recordings with one, three, six, or twelve leads. The system used in this study directly records two leads, from which an additional four leads are subsequently calculated.

%%%%%%%%%%%%%%%%%%%%%
\subsection{Noise in ECG signals}
%%%%%%%%%%%%%%%%%%%%%

Although ECGs are intended to provide a true and accurate representation of myocardial electrical activity, the recorded signals are inherently vulnerable to various forms of electrical interference, collectively referred to as \textbf{noise}. Such interference can significantly distort waveform morphology and alter temporal intervals, thereby impairing the ability of veterinarians, cardiologists, and automated interpretation algorithms to reliably detect, measure, and classify cardiac events or identify pathological abnormalities.

Sources of electrical interference can be broadly categorized as patient-related or environmental. Patient-related noise commonly includes respiratory artifacts that manifest as baseline wander, as well as electromyographic interference arising from muscle activity. Environmental noise sources include poor electrode-skin contact and electromagnetic interference from nearby electrical devices or even distant power lines. In recorded ECG signals, these disturbances typically appear as superimposed extraneous waveforms that obscure clinically relevant cardiac features.

%%%%%%%%%%%%%%%%%%%%%
\subsection{Canine ECG characteristics and interpretation}
%%%%%%%%%%%%%%%%%%%%%

In canine electrocardiography, specific leads provide characteristic views of the cardiac cycle.
\begin{itemize}
    \item \textbf{Lead I} records the electrical activity predominantly in the frontal plane, reflecting current flow from the right forelimb to the left forelimb.
    \item \textbf{Lead II} captures the current flow from the right forelimb to the left hindlimb. Due to the more vertical orientation of the canine heart within the thoracic cavity compared to humans, lead II often provides the clearest visualization of the $P-QRS-T$ sequence.
\end{itemize}

The key components of the canine ECG waveform include:
\begin{itemize}
    \item The \textbf{$P$-wave}, representing atrial depolarization, is the first positive deflection in a typical cardiac cycle and is usually upright in lead II. Its consistent presence and uniform morphology across cycles suggest an origin from the sinoatrial (SA) node. In lead I, the $P$-wave may be upright or biphasic, influenced by the individual animal's cardiac axis.
    \item The \textbf{$QRS$-complex}, corresponding to ventricular depolarization, follows the $P$-wave. In many dogs, this complex appears primarily negative in lead I and can be positive or isoelectric in lead II, depending on heart orientation and body conformation. A narrow and sharply defined $QRS$-waveform indicates efficient conduction through the His-Purkinje system.
    
    \item The \textbf{$T$-wave}, reflecting ventricular repolarization, follows the $QRS$-complex. Its polarity can be variable (positive, negative, or biphasic) in lead II without necessarily indicating pathology, particularly in breeds with diverse thoracic conformations.
\end{itemize}

%%%%%%%%%%%%%%%%%%%%%
\subsection{Normal sinus rhythm in dogs}
%%%%%%%%%%%%%%%%%%%%%
Normal sinus rhythm in dogs is defined by a predictable and orderly sequence of these electrical events. Key features, prominently observed in leads I and II, include:
\begin{itemize}
    \item A $P$-wave preceding every $QRS$-complex, confirming organized atrial activation.
    \item Upright $P$-waves in lead II, consistent with an SA nodal origin of the impulse.
    \item Consistent $P-QRS-T$ wave sequences, repeated uniformly throughout the tracing.
    \item Regular $R-R$ intervals, indicating a steady heart rate without premature or delayed beats.
\end{itemize}

The consistent presence of these characteristics in leads I and II confirms that electrical impulses are initiated in the SA node and conducted appropriately through the atrioventricular (AV) node and ventricles. While anatomical and breed-specific variations can lead to some waveform differences, the rhythm's consistency and overall pattern are the primary indicators of a sinus origin. Consequently, in canine ECG interpretation, leads I and II are often sufficient for confirming normal sinus rhythm and establishing a baseline for further cardiac assessment.

%%%%%%%%%%%%%%%%%%%%%%%%%%%%%%%%%%%%%%%%%%
\section{Materials and methods}
%%%%%%%%%%%%%%%%%%%%%%%%%%%%%%%%%%%%%%%%%%

%%%%%%%%%%%%%%%%%%%%%
\subsection{Dataset acquisition and preparation}
%%%%%%%%%%%%%%%%%%%%%

A primary challenge in applying machine learning to ECG denoising lies in the acquisition and preparation of a suitable dataset. This process involves several critical considerations:

\begin{itemize}
    \item \textbf{Data availability and clean signal definition}: Obtaining truly noise-free ECG recordings from canines in real-world clinical settings is often infeasible due to their naturally high levels of movement and activity. As a result, acquiring a sufficient volume of ECG segments that can be considered for model training is itself a resource-intensive process. For this study, we meticulously curated a dataset of $10,000$ ten-second ECG segments that met this criterion, which served as clean reference signals. All ECG recordings were annotated by a single DVM annotator, who classified segments as clean based solely on the minimal presence of common noise artifacts, including baseline wander, electrical interference, and muscle interference.

    \item \textbf{Generation of realistic noisy data}: Training a denoising model requires paired examples of noisy ECG signals and their corresponding clean reference signals. However, the nature of noise in ECG data is highly heterogeneous, encompassing both stochastic perturbations and structured, often periodic interference. Noise may contaminate an entire recording or be confined to brief temporal segments, and its manifestation can vary across individual leads, with some channels exhibiting pronounced artifacts while others remain relatively unaffected. Furthermore, multiple distinct noise sources may coexist within a single ECG recording. Faithfully capturing this complex and multifaceted noise behavior is essential for developing a robust and generalizable denoising model. Accordingly, our approach involved generating synthetic noisy ECG signals from the curated set of clean reference segments, thereby constructing a controlled and representative training dataset.

    \item \textbf{Model generalization and role of clean-clean pairs}: For a denoising model to be practically useful, it must generalize effectively to previously unseen noisy signals while also preserving signal integrity when presented with signals that are already clean. Achieving these dual objectives requires a diverse and representative training dataset. We investigated the inclusion of (clean, clean) signal pairs within the training data. These pairs provide the model with an explicit reference for the characteristics of clean ECG signals, anchoring the learning process and reinforcing an identity mapping when no noise is present. The (clean, clean) pairs also facilitate evaluation of the model's ability to avoid unnecessary signal alteration and play a critical role in validation and testing by enabling direct assessment of output fidelity. We systematically examined different proportions of such pairs to determine the configuration most conducive to optimal overall model performance.

\end{itemize}

%%%%%%%%%%%%%%%%%%%%%
\subsection{Defining noise transforms}
%%%%%%%%%%%%%%%%%%%%%

Defining sufficiently complex and stochastic noise transformations is foundational to the development of an effective ECG denoising model. We designed a set of noise transformations applied to the curated dataset of ECG signals that emulate the types of interference commonly encountered in real-world ECG recordings. To reflect the fact that noise may affect individual leads differently, these transformations are randomly applied to a subset of leads within each multilead recording, while other leads may remain relatively clean. This strategy captures realistic, lead-specific noise behavior and enables the model to exploit complementary electrical perspectives across multiple leads during denoising, thereby enhancing its ability to recover the underlying physiological signal. The introduction of controlled yet diverse noise patterns during training allows the model to learn representative noise characteristics while preserving essential cardiac features.

%%%%%%%%%%%%%%%%%%%%%
\subsubsection{Sine baseline wander}
\begin{figure}
    \centering
    \includegraphics[width=0.75\linewidth]{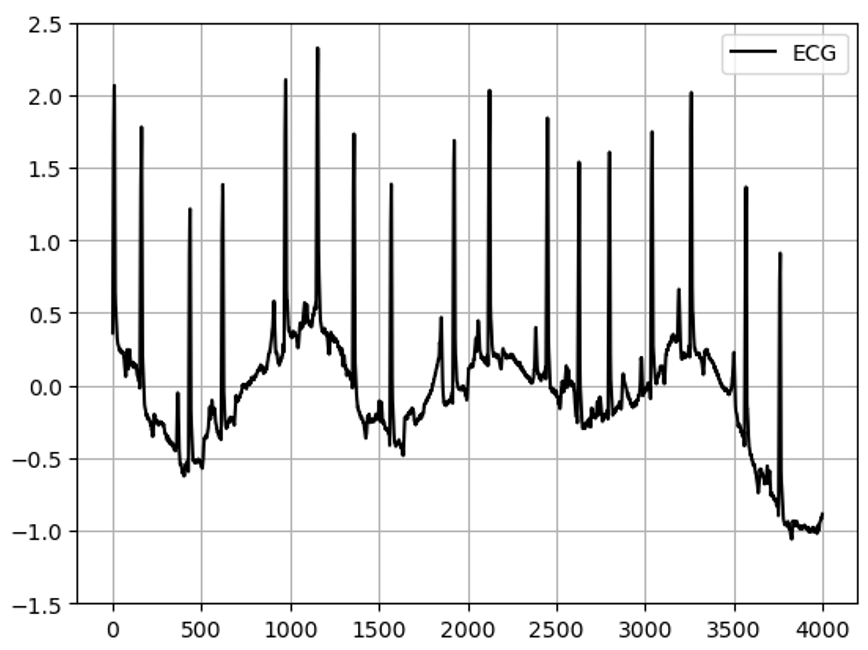}
    \caption{An ECG lead with baseline wander.}
    \label{fig:ecgleadbaselinewander}
\end{figure}

%%%%%%%%%%%%%%%%%%%%%
In the development process, we begin by defining {\em sine baseline wander} transformations, which represent a common form of noise in ECG recordings. Baseline wander is typically caused by low-frequency interference associated with respiration-induced motion. The effect of this interference is well approximated by the superposition of a sinusoidal waveform on the underlying ECG signal, as illustrated in Figure \ref{fig:ecgleadbaselinewander}. In this figure, as in all ECG plots presented in this paper, signals are sampled at a rate of $500$ samples per second; consequently, the horizontal axis denotes the sample index, with index $500$ corresponding to one second of recorded time. The vertical axis represents the measured electrical potential in millivolts (mV).

As with other forms of ECG interference, baseline wander may affect an entire recording or be confined to localized temporal segments. Furthermore, sine baseline wander may be present in only a subset of leads rather than uniformly across all leads, reflecting the lead-specific and transient nature of respiratory-related artifacts observed in real-world ECG data.

\begin{figure}
    \begin{center}
    \begin{adjustbox}{valign=c}
    \includegraphics[width=0.3\linewidth]{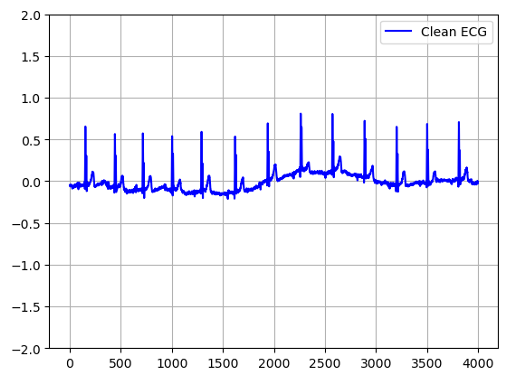}
    \end{adjustbox}
    \begin{adjustbox}{valign=c}
    $+$
    \end{adjustbox}
    \begin{adjustbox}{valign=c}
    \includegraphics[width=0.3\linewidth]{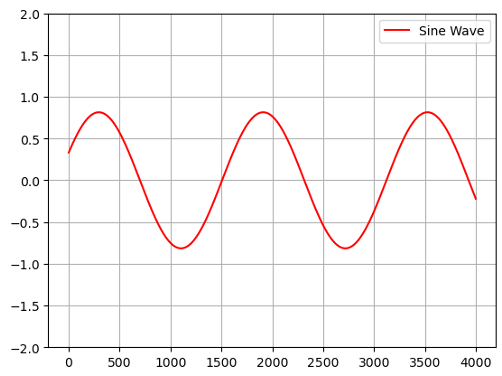}
    \end{adjustbox}
    \begin{adjustbox}{valign=c}
    $=$
    \end{adjustbox}
    \begin{adjustbox}{valign=c}
    \includegraphics[width=0.3\linewidth]{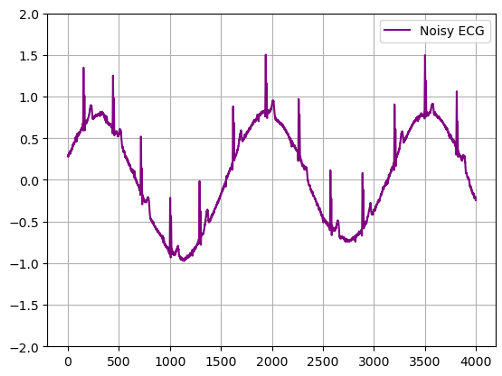}
    \end{adjustbox}
    \caption{Baseline wander noising process.}
    \label{fig:cleanplusbaselinewander}
    \end{center}
\end{figure}

To mimic baseline wander, we synthetically add a sinusoidal component to an ECG segment, as shown in Figure \ref{fig:cleanplusbaselinewander}. Each baseline wander realization for a given lead is defined by the following parameters:
\begin{itemize}
    \item the start and end points of the affected segment,
    \item the amplitude,
    \item the frequency, and
    \item the phase shift.
\end{itemize}

For each added sinusoidal component, these parameters are randomly sampled from a predefined search space. To emulate baseline wander observed in real-world ECG recordings, the amplitude is drawn from a narrow range informed by empirical ECG data, while the frequency is selected from a range consistent with expected respiratory rates. To ensure a sufficient representation of baseline wander within each ECG segment, the sinusoidal component, when selected, is applied to at least $60\%$ of the signal duration, allowing controlled variability in both the extent and severity of baseline wander across training examples.

%%%%%%%%%%%%%%%%%%%%%
\subsubsection{White noise}
%%%%%%%%%%%%%%%%%%%%%
Muscle activity generates electrical signals that can introduce distortions into the ECG recording. Involuntary movements, such as shivering or muscle tremors, as well as voluntary actions like repositioning, often produce high-frequency artifacts that appear as small, irregular fluctuations superimposed on the ECG signal. Although measures such as improving patient comfort and relaxation can reduce the prevalence of this type of noise, complete elimination is rarely achievable in practice. Muscle-related interference can obscure clinically relevant features, including low-amplitude waves such as the $P$-wave or $T$-wave, and in some cases may mimic arrhythmias such as atrial flutter.

Similar high-frequency disturbances may also arise from electronic components within the ECG acquisition system or from external sources of electromagnetic interference, including nearby electrical devices or power lines. In this work, we collectively refer to noise arising from muscle activity and external electrical interference as {\em white noise}, reflecting its broadband, irregular characteristics.

Each white noise component is parameterized by
\begin{itemize}
    \item the start and end points of the affected segment,
    \item the amplitude, and
    \item the frequency.
\end{itemize}

To ensure adequate representation of this noise type within the training data, white noise, when selected, is added to at least $20\%$ of each ECG segment, allowing for controlled variability in both the temporal extent and intensity of the interference.

%%%%%%%%%%%%%%%%%%%%%
\subsubsection{Linear baseline wander}
%%%%%%%%%%%%%%%%%%%%%
We also observed the presence of {\em linear baseline wander} in ECG recordings, which may arise from patient movement, respiratory motion, or inadequate electrode-skin contact. Unlike sinusoidal baseline wander, this form of interference manifests as a gradual, approximately linear drift in the ECG baseline over time.

Each instance of linear baseline wander applied to a given lead is defined by the following parameters:
\begin{itemize}
    \item the start and end points of the affected segment, and
    \item the slope of the baseline drift.
\end{itemize}

To ensure adequate representation of linear baseline drift within the training data, this noise transformation, when selected, is applied to at least $60\%$ of the ECG segment, allowing variation in both the duration and magnitude of the baseline shift.

%%%%%%%%%%%%%%%%%%%%%
\subsubsection{Shock pulses}
%%%%%%%%%%%%%%%%%%%%%
{\em Shock pulses} in ECGs can be caused by sudden extreme movements or poor lead connections. We define the shock pulses using a restricted sine wave.

The parameters that define the specific shock pulses for a lead are:
\begin{itemize}
    \item the number of pulses,
    \item the frequency of the pulse, and,
    \item the maximum value of the pulse.
\end{itemize}

%%%%%%%%%%%%%%%%%%%%%
\subsubsection{Multilead and multi-noise data augmentation}
%%%%%%%%%%%%%%%%%%%%%
Each $({\rm noisy\, ECG}, {\rm clean\, ECG})$ pair in the training, validation and testing PyTorch datasets is constructed using a randomized, multi-stage noise application strategy. First, a random subset of leads is selected for noising, reflecting the fact that noise does not uniformly affect all leads in real-world recordings. For each selected lead, a random subset of noise types is then chosen from the predefined noise families. For each selected noise type, the parameters governing the corresponding noise transformation are independently and randomly sampled subject to predefined constraints. These noise transformations are subsequently applied to the clean ECG signals to generate the corresponding noised ECGs for each dataset pair that is not of the $({\rm clean}, {\rm clean})$ type.

To prevent data leakage, ECG segments included in $({\rm clean}, {\rm clean})$ pairs are explicitly excluded from participation in any $({\rm noised\, ECG}, {\rm clean\, ECG})$ pairs. This ensures that clean reference signals used for identity-mapping or validation purposes are not reused as targets for denoising during training. Overall, this procedure enables the generation of a large, diverse, and realistic set of noisy ECG recordings, supporting the development of robust denoising models capable of handling complex and co-occurring distortions across leads. For performance evaluation and reproducibility, the random state governing noise generation is fixed for the validation and testing datasets.

%%%%%%%%%%%%%%%%%%%%%
\subsection{Denoising models}
%%%%%%%%%%%%%%%%%%%%%

We conducted experiments using both \textbf{autoencoder} and \textbf{diffusion} model architectures for the ECG denoising task. These model classes were selected for their well-established effectiveness in signal reconstruction and generative modeling, particularly in the context of ECG denoising and restoration \cite{badiger2023ascnetecg}, \cite{cardoso2023bayesianecgreconstructionusing}, \cite{chiang2019noisereduction}, \cite{ho2020denoisingdiffusionprobabilisticmodels}, \cite{li2024descodecg}, \cite{romero2021deepfilterecgbaselinewander}, \cite{xiong2015denoising}, \cite{xu2021ecgdenoising}. While both approaches demonstrated strong denoising performance in our experiments, we focused primarily on autoencoder-based models due to their substantially faster training times. This computational efficiency enabled more extensive experimentation with different noise configurations and architectural variants without compromising overall model performance. The final autoencoder architecture used for the denoising and delineation results presented in this paper operates on ECG leads I and II simultaneously. It consists of $6$ encoder layers with hidden channel sizes $(16, 32, 64, 128, 256, 512)$, and $6$ decoder layers with sizes $(512, 256, 128, 64, 32, 16)$. All convolutional layers use a kernel size of 16, with stride 1 and zero padding throughout.

%%%%%%%%%%%%%%%%%%%%%
\subsection{Evaluation strategy}
%%%%%%%%%%%%%%%%%%%%%

The performance of the denoising models was evaluated using a multi-faceted strategy designed to assess both signal fidelity and downstream clinical utility:
\begin{itemize}
    \item \textbf{Signal similarity metrics}: Quantitative similarity metrics were used to measure the agreement between the denoised ECG signals and their corresponding reference signals, providing an objective assessment of denoising quality at the waveform level.
    \item \textbf{Downstream task performance}: The practical and clinical relevance of the denoised signals was evaluated by measuring improvements in performance on downstream ECG delineation tasks, including the identification of $P$-waves, $QRS$ complexes, and $T$-waves.
\end{itemize}

%%%%%%%%%%%%%%%%%%%%%
\subsubsection{Signal similarity metrics}
%%%%%%%%%%%%%%%%%%%%%

Signal similarity metrics aim to quantify the extent to which noise is introduced into or removed from an ECG signal, thereby providing a direct measure of denoising effectiveness. These metrics evaluate how closely a denoised signal matches its corresponding reference signal and offer insight into the preservation of underlying physiological features. The measures employed in this study are standard within the signal processing literature and are commonly used for assessing ECG noise reduction performance. In particular, metrics such as the signal-to-noise ratio (SNR) provide an interpretable summary of the relative strength of the desired cardiac signal compared to residual noise. The following metrics are used to quantify noise and to compare two finite-length signals of the same size, denoted $X$ and $Y$.

\begin{defn}
Let $X=[x_1, x_2, \dots, x_N]$ be a finite sequence of real values $x_i$. The {\em signal power} of $X$ is defined as
$${\rm SignalPower}(X)=\sum_{i=1}^Nx_i^2.$$
Additionally, when there exists $1\leq i\leq N$ with $x_i \neq 0$, the {\em logarithmic scaled signal power} of $X$ is defined as
$${\rm SignalPower}_{\log}(X)=10\cdot\log_{10}\sum_{i=1}^N x_i^2.$$
\end{defn}

\begin{defn}
Let $X=[x_1, x_2, \dots, x_N]$ and $\xi=[\xi_1, \xi_2, \dots, \xi_N]$ be finite sequences of real values of the same length such that there exists $1\leq i\leq N$ with $x_i \neq 0$ and there exists $1\leq j\leq N$ with $x_j+\xi_j\neq 0$ . The {\em signal-to-noise ratio}, or {\em SNR}, of $X$ to $X+\xi$ is defined as
\begin{eqnarray*}
{\rm SNR}(X, X+\xi) &=& {\rm SignalPower}_{\log}(X) - {\rm SignalPower}_{\log}(X+\xi)\\
&=&10\cdot\log_{10}\left(\dfrac{{\rm SignalPower}(X)}{{\rm SignalPower}(X+\xi)}\right)
\end{eqnarray*}
\end{defn}

\begin{defn}
Let $X=[x_1, x_2, \dots, x_N]$ and $Y=[y_1, y_2, \dots, y_N]$ be finite sequences of real values of the same length. The {\em sum of squared differences}, or {\em SSD}, of $X$ and $Y$ is defined as
$${\rm SSD}(X, Y) = \sum_{i=1}^N (x_i - y_i)^2.$$
\end{defn}

\begin{defn}
Let $X=[x_1, x_2, \dots, x_N]$ and $Y=[y_1, y_2, \dots, y_N]$ be finite non-zero sequences of real values of the same length. The {\em cosine similarity} of $X$ and $Y$ is defined as the cosine of the angle $\theta$ between $X$ and $Y$. That is,
$$\cos\theta = \dfrac{X\cdot Y}{||X||\cdot||Y||}.$$
We also define the {\em cosine distance} of $X$ and $Y$ to be $1$ minus the cosine similarity of $X$ and $Y$.
\end{defn}

%%%%%%%%%%%%%%%%%%%%%
\subsubsection{Delineation-based evaluation metrics}\label{delineation_metrics}
%%%%%%%%%%%%%%%%%%%%%
The second class of evaluation metrics focuses on quantifying the impact on the performance of ECG delineation models. The primary objective of this evaluation is to assess how noise degrades delineation accuracy and to measure the extent to which denoising improves downstream delineation performance. These metrics are inspired by object detection and classification frameworks commonly used in computer vision, adapted to the structured and temporal nature of ECG signals.

Specifically, the delineation task aims to (1) correctly detect true cardiac electrical waves, (2) accurately classify detected waves as $P$-, $QRS$-, or $T$-waves, and (3) determine whether each identified wave is normal or abnormal. Identified waves are characterized by their peak locations and temporal extents (i.e., start and end points). Accurate delineation is essential for computing clinically relevant derived measures, such as $RR$-intervals and $ST$-segment durations. Minor localization discrepancies are tolerated, provided they do not affect overall delineation performance or downstream clinical interpretation.

ECG signals possess a well-defined and repetitive structure, making precise identification of $P$-, $QRS$-, and $T$-waves fundamental. Accordingly, delineation performance is evaluated using counts of true positives, false positives, and false negatives for wave detection and classification. Additional metrics include peak localization accuracy, interval overlap accuracy, and confusion matrices for normal versus abnormal wave classification.

The delineation model outputs, for each ECG segment, include the predicted wave type ($P$, $QRS$, or $T$), the peak index, the wave interval (starting and ending indices), and a binary normal/abnormal classification for each detected wave.

In the {\bf wave matching procedure}, predicted delineation outputs are first aligned with ground truth annotations. For each wave type ($P$, $QRS$, and $T$), a ground truth peak index is matched to the closest predicted peak index of the same wave type, provided the distance between them falls within a predefined matching threshold. The absolute difference between the matched predicted peak index and the corresponding ground truth peak index is then recorded.

For {\bf peak classification}, a precision threshold of 2 milliseconds is applied to peak localization. If a matched predicted peak lies within this threshold of the corresponding ground truth peak, it is classified as a {\em true positive}. If a predicted peak does not have a corresponding ground truth peak within the threshold, it is classified as a {\em false positive}. Conversely, if a ground truth peak has no associated predicted peak within the threshold, it is classified as a {\em false negative}. True negative classifications are not considered, as wave peaks are sparse events in ECG signals and inclusion of this category would not be informative.

We evaluate wave intervals using {\bf interval-based metrics} computed for matched waves by comparing the predicted wave intervals with their corresponding ground truth intervals. Specifically, the length of the intersection between the predicted and ground truth intervals is computed, and the degree of overlap is quantified as a proportion of each interval. For example, if the predicted wave fully contains the ground truth wave, the intersection is equal in length to the ground truth wave, yielding a $100\%$ overlap relative to the ground truth interval:
\[
\frac{|{\rm predicted\ interval} \cap {\rm ground\ truth\ interval}|}{|{\rm predicted\ interval}|}
=
\frac{|{\rm ground\ truth\ interval}|}{|{\rm predicted\ interval}|}.
\]
However, there may be a potentially small overlap relative to the predicted interval.

An analogous situation arises when the ground truth interval fully contains the predicted interval, in which case the overlap represents a $100\%$ share of the predicted interval but a potentially small share of the ground truth interval.

A wave interval prediction is classified as a {\em true positive} if the intersection constitutes at least $80\%$ of both the predicted interval and the ground truth interval. If a predicted interval lacks a valid corresponding ground truth interval or fails to meet this overlap threshold, it is classified as a {\em false positive}. Conversely, if a ground truth interval lacks a corresponding predicted interval meeting the overlap criterion, it is classified as a {\em false negative}. As with peak detection, true negative classifications are not considered for interval evaluation, as wave events are sparse within ECG signals and this category would not be informative.

For matched waves, the predicted {\bf normal/abnormal classification} is compared against the ground truth label. A prediction is classified as a {\em true positive} if both the predicted wave and the ground truth wave are abnormal, and as a {\em true negative} if both are normal. A {\em false positive} occurs when a normal ground truth wave is predicted as abnormal, while a {\em false negative} occurs when an abnormal ground truth wave is predicted as normal.

Finally, {\bf composite metrics} are computed by combining peak detection, interval accuracy, and normal/abnormal classification results. A predicted wave is classified as a {\em true positive peak-wave} if both its peak and interval are correctly identified. If either the peak or the interval is a false positive, the peak-wave is classified as a {\em false positive}. If either component is a false negative, the peak-wave is classified as a {\em false negative}. {\bf Complete agreement} is achieved when a predicted wave satisfies peak accuracy, interval overlap, and correct normal/abnormal classification simultaneously.

%%%%%%%%%%%%%%%%%%%%%%%%%%%%%%%%%%%%%%%%%%
\section{Results}
%%%%%%%%%%%%%%%%%%%%%%%%%%%%%%%%%%%%%%%%%%

In the denoising-delineation ECG pipeline
\begin{center}
noisy ECG $\to$ denoised ECG $\to$ delineated ECG,
\end{center}
there are two natural points at which denoising model performance can be evaluated: immediately after denoising and after downstream ECG delineation. At the first evaluation point, we quantify changes in signal quality using signal-to-noise ratio (SNR) and the additional noise metrics described above. As baselines, we compare the performance of our deep learning models against classical signal filtering techniques.

At the second evaluation point, we assess performance using the delineation metrics introduced previously. Because the primary goal of denoising is to improve clinically relevant interpretation rather than purely aesthetic signal quality, delineation performance is prioritized over waveform-level noise metrics. Notably, we observe that classical filters may occasionally achieve higher scores on denoising metrics; however, these improvements do not consistently translate into better delineation performance. In contrast, ECGs processed by our deep learning denoising models frequently yield superior delineation results despite appearing less visually smooth.

To illustrate these effects, we present a representative example. In all delineation visualizations,  \textcolor{red}{$P$-waves are shown in red}, \textcolor{darkgreen}{$QRS$-complexes are shown in green}, and \textcolor{blue}{$T$-waves are shown in blue}.

\begin{figure}
    \centering
    \includegraphics[width=1\linewidth]{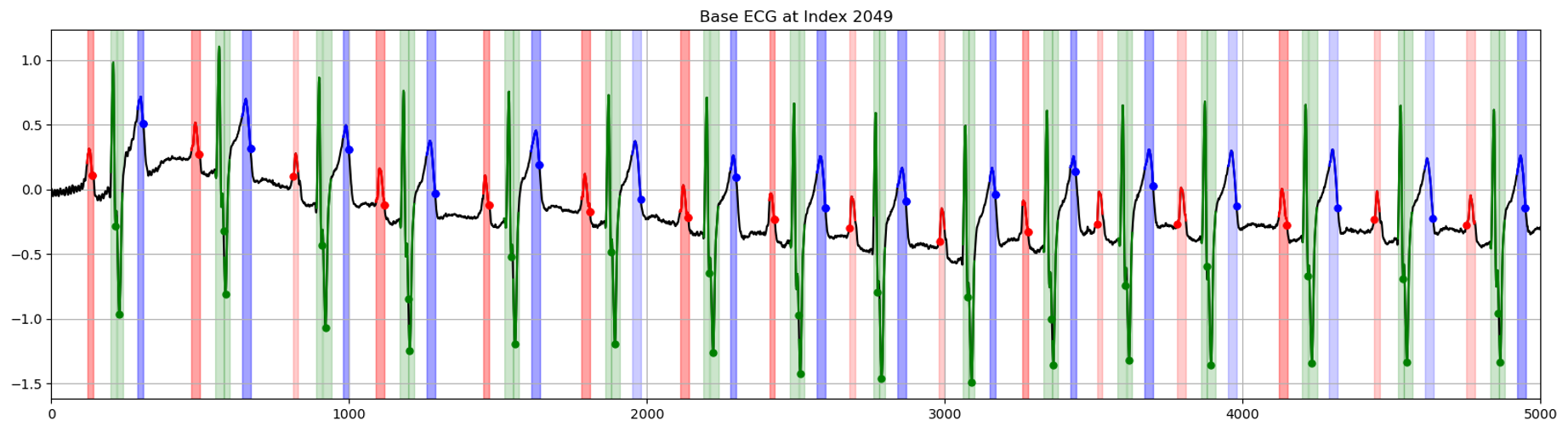}
    \caption{A delineated base ECG.}
    \label{fig:basedelineation}
\end{figure}

When an ECG is sufficiently clean, the delineation model accurately identifies $P$-waves, $QRS$-complexes, and $T$-waves and correctly classifies them as normal or abnormal, as shown in Figure \ref{fig:basedelineation}.

\begin{figure}
    \centering
    \includegraphics[width=1\linewidth]{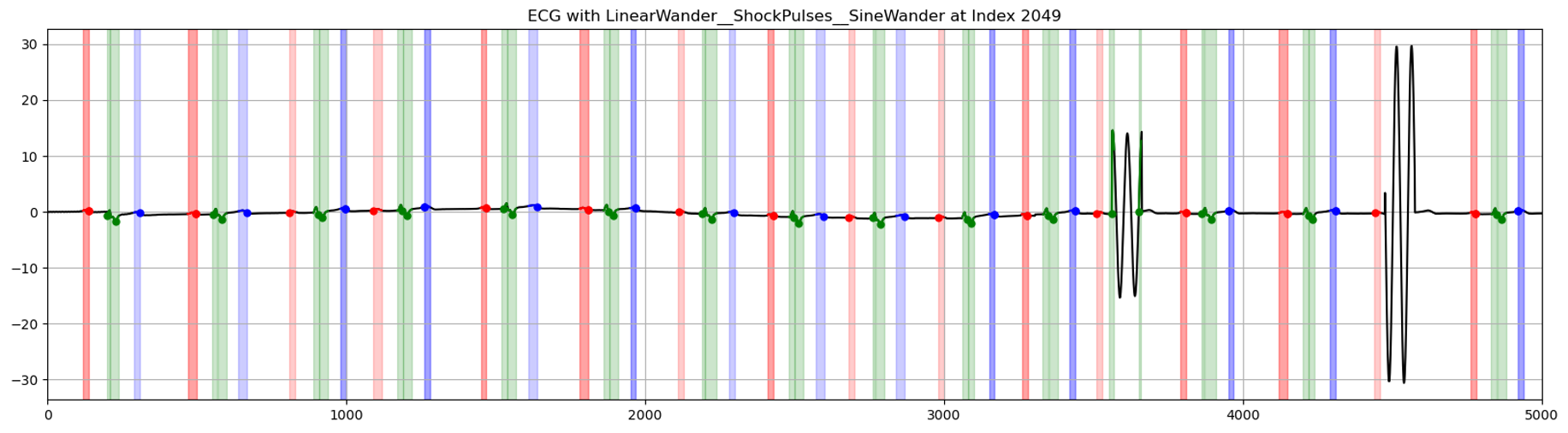}
    \caption{The delineated noised base ECG.}
    \label{fig:noiseddelineation}
\end{figure}

When the same ECG is corrupted by noise, delineation performance degrades substantially. Figure \ref{fig:noiseddelineation} shows the delineation output for the ECG in Figure \ref{fig:basedelineation} after the addition of linear baseline wander, shock pulses, and sinusoidal baseline wander. The injected noise results in $187$ delineation errors across wave classification, peak localization, wave interval detection, and normal/abnormal classification.

\begin{figure}
    \centering
    \includegraphics[width=1\linewidth]{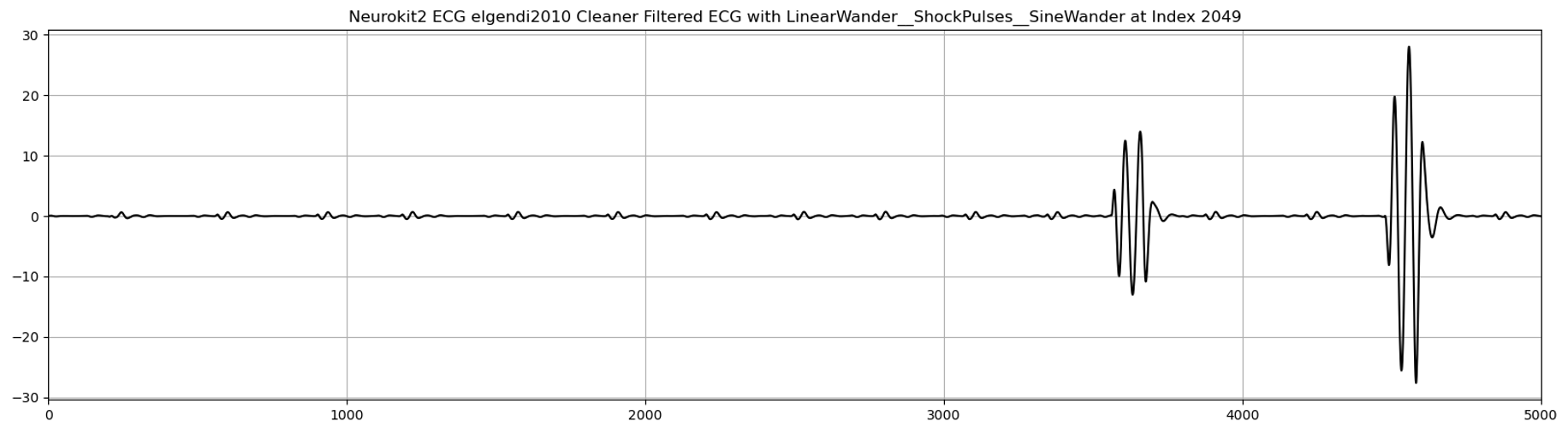}
    \caption{The elgendi2010 filtered noised ECG.}
    \label{fig:filteredecg}
\end{figure}

Applying the classical \verb"elgendi2010" filter available from the \verb"Neurokit2" package to the noised ECG produces the filtered signal shown in Figure \ref{fig:filteredecg}.

\begin{figure}
    \centering
    \includegraphics[width=1\linewidth]{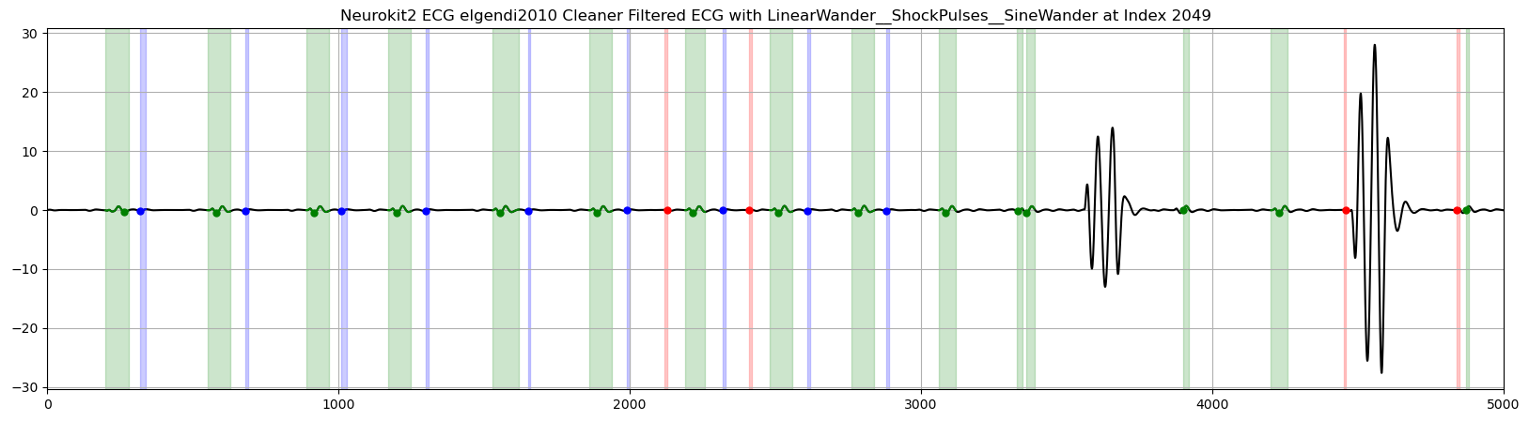}
    \caption{The delineated elgendi2010 filtered noised ECG.}
    \label{fig:delineatedfilteredecg}
\end{figure}

When the delineation model is subsequently applied to this filtered ECG, the resulting output (Figure \ref{fig:delineatedfilteredecg}) exhibits a further degradation in delineation performance. Specifically, $119$ additional delineation errors are introduced, yielding a total of $306$ errors across wave classification, peak identification, interval detection, and normal/abnormal labeling.

\begin{figure}
    \centering
    \includegraphics[width=1\linewidth]{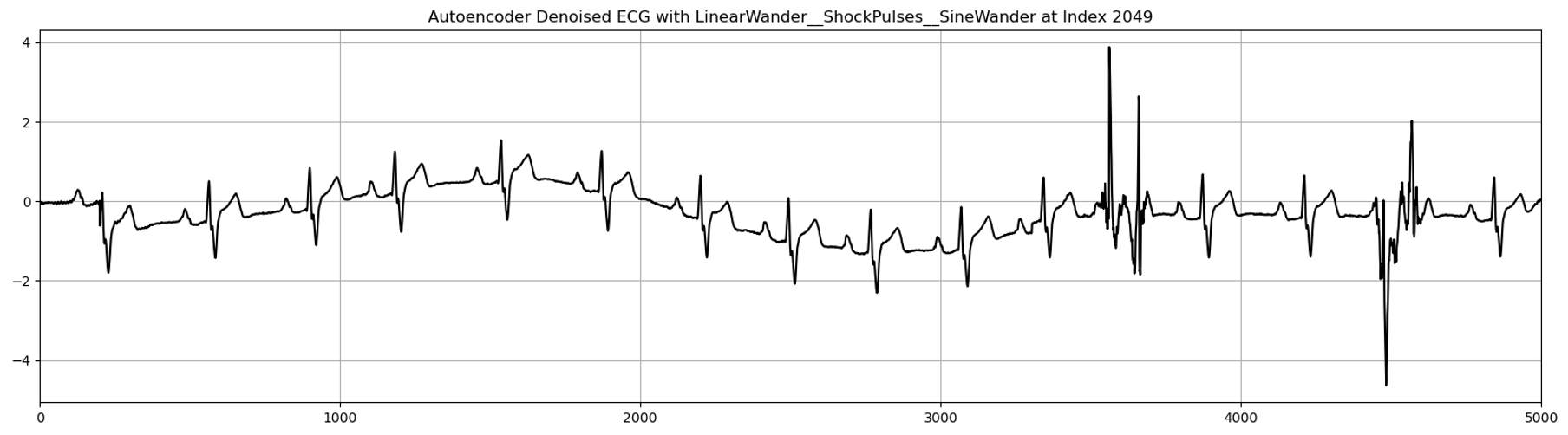}
    \caption{The autoencoder cleaned noised ECG.}
    \label{fig:denoisedecg}
\end{figure}

In contrast, applying our autoencoder-based denoising model to the noised ECG produces the signal shown in Figure \ref{fig:denoisedecg}. Although the autoencoder-denoised ECG does not appear as visually smooth as the filtered signal, it yields markedly improved delineation performance. In this example, the autoencoder corrects $77$ of the $187$ delineation errors introduced by the added noise.

\begin{figure}
    \centering
    \includegraphics[width=1\linewidth]{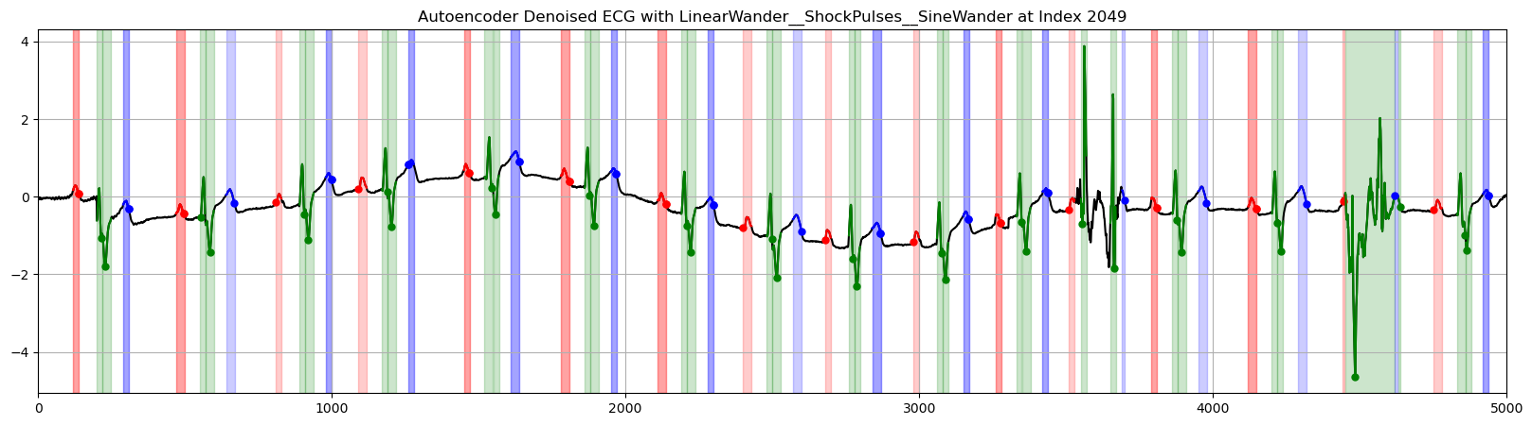}
    \caption{The delineated autoencoder cleaned noised ECG.}
    \label{fig:delineateddenoisedecg}
\end{figure}

The delineation output for the autoencoder-denoised ECG is shown in Figure \ref{fig:delineateddenoisedecg}. Compared with both the filtered ECG (Figure \ref{fig:delineatedfilteredecg}) and the original noised ECG (Figure \ref{fig:noiseddelineation}), the autoencoder-denoised signal demonstrates substantially improved identification of wave peaks, intervals, and normal/abnormal classifications. Relative to the baseline delineation in Figure \ref{fig:basedelineation}, these visual comparisons confirm the superiority of our autoencoder-based denoising approach over the selected classical filter for improving downstream ECG delineation.

%%%%%%%%%%%%%%%%%%%%%
\subsection{Denoising performance metrics}
%%%%%%%%%%%%%%%%%%%%%

Across both noisy and clean ECG signals, our autoencoder-based denoising model consistently produces ECG inputs that remain suitable for downstream delineation. While certain classical filtering techniques may slightly outperform the autoencoder for specific, isolated noise types, they generally struggle to handle diverse noise patterns and clean signals simultaneously. In contrast, autoencoder architectures can be scaled through increased model capacity and can continually improve as additional training data become available, overcoming limitations inherent to fixed-parameter traditional signal filters.

To evaluate denoising performance, we compare classical filters and our autoencoder on a dataset of $5,000$ ten-second ECG segments, both without applied noise and with synthetic noise added according to the procedures described above. Performance is assessed at two stages: immediately following denoising and after downstream delineation.

For the selected reference ECG signals, we compute the sum of squared differences ($SSD$), maximum absolute difference ($MAD$), and cosine distance between the denoised signals and their original reference signals for each signal-cleaning method. For all three metrics, values closer to zero indicate better denoising performance. Results for clean signals are summarized in Tables \ref{table:cleaned_SSD}, \ref{table:cleaned_MAD}, and \ref{table:cleaned_COSDIST}.

For ECGs with added noise, the magnitude of injected noise varies across segments. To enable meaningful comparisons, we report the proportion of added noise removed by each signal-cleaning method, measured in terms of $SSD$, $MAD$, and cosine distance. For example, the percentage of noise removed according to $SSD$ is computed as
$$\dfrac{added\_SSD - cleaned\_SSD}{added\_SSD}.$$
Positive values indicate successful noise reduction, whereas negative values indicate that the denoising method introduced additional distortion relative to the noisy input.

We report these noise-removal metrics for the following noise configurations:
\begin{itemize}
    \item \verb"LinearWander",
    \item \verb"SineWander", and
    \item \verb"LinearWander-MuscleArtifact-ShockPulses-SineWander".
\end{itemize} 

Results for \verb"LinearWander" are reported in Tables \ref{table:LinearWandercleaned_SSD}, \ref{table:LinearWandercleaned_MAD}, and \ref{table:LinearWandercleaned_COSDIST}. Metrics for \verb"SineWander" are reported in Tables \ref{table:SineWandercleaned_SSD}, \ref{table:SineWandercleaned_MAD}, and \ref{table:SineWandercleaned_COSDIST}. Finally, metrics for the combined multi-noise condition \\ \verb"LinearWander-MuscleArtifact-ShockPulses-SineWander" are reported in Tables  \ref{table:multinoisecleaned_SSD}, \ref{table:multinoisecleaned_MAD}, and \ref{table:multinoisecleaned_COSDIST}.

%%%%%%%%%%%%%%%%%%%%%
\subsection{Delineation performance metrics}
%%%%%%%%%%%%%%%%%%%%%

Across both clean and noisy ECG signals, our autoencoder‑based denoising model consistently matches or outperforms traditional filtering approaches with respect to downstream wave delineation performance. When evaluated on the curated reference ECG signals, the autoencoder achieves delineation accuracy comparable to, and in several cases exceeding, that of the best-performing classical filter, as shown in Table \ref{table:denoised_clean_delineation}.

A similar trend is observed for ECGs corrupted by synthetic noise. As reported in Tables \ref{table:delineated_LinearWander}, \ref{table:delineated_SineWander}, and \ref{table:delineated_multinoise}, the autoencoder-based approach consistently achieves delineation performance that is equal to or superior to that of traditional filters across a range of noise types and combinations.

The ability of the proposed denoising model to maintain strong delineation performance under both clean and noisy conditions highlights its robustness and adaptability. These results demonstrate that optimizing denoising specifically for downstream delineation, rather than solely for waveform smoothness or noise suppression, yields more reliable and clinically relevant interpretations across diverse signal conditions.

%%%%%%%%%%%%%%%%%%%%%%%%%%%%%%%%%%%%%%%%%%
\section{Conclusion}
%%%%%%%%%%%%%%%%%%%%%%%%%%%%%%%%%%%%%%%%%%

In this work, we demonstrated that a deep learning-based denoising approach can consistently match or outperform classical signal filtering techniques for ECG processing while offering substantially greater flexibility for future improvement. Although traditional signal processing methods can be parameterized, they are not learning-based and therefore cannot improve with additional data or readily adapt to new, complex noise patterns while simultaneously preserving already clean ECG signals.  

In contrast, the autoencoder architectures proposed in this study can be continuously refined through exposure to larger and more diverse datasets or by increasing model capacity via deeper or wider network designs. This adaptability enables the model to better capture the complex and heterogeneous noise characteristics encountered in real-world ECG recordings. As evidenced by our experiments, classical filters often degrade downstream delineation performance by redistributing noise across the signal rather than selectively removing it, even when exhibiting favorable waveform-level denoising metrics.

By explicitly optimizing denoising for downstream delineation performance, our approach produces ECG signals that are more suitable for accurate wave identification, interval estimation, and clinical interpretation. These results highlight the importance of evaluating denoising techniques not only by signal-level noise metrics but also by their impact on clinically relevant tasks. Overall, the proposed autoencoder-based framework presents a robust, extensible, and clinically meaningful alternative to traditional ECG denoising methods.

%%%%%%%%%%%%%%%%%%%%%%%%%%%%%%%%%%%%%%%%%%
\section{Acknowledgments}
%%%%%%%%%%%%%%%%%%%%%%%%%%%%%%%%%%%%%%%%%%
The authors express their sincere gratitude to Andre Dourson, Michael Fitzke, Mark Parkinson, Xiaoli Qiao, and Kylie Taylor for their invaluable support throughout this work. Their thoughtful feedback on earlier drafts, constructive discussions, and generous sharing of expertise substantially improved both the rigor and clarity of the analysis presented in this paper. The authors are deeply grateful for the time and insight each of them contributed.

%%%%%%%%%%%%%%%%%%%%%%%%%%%%%%%%%%%%%%%%%%%%%%%
% References
%%%%%%%%%%%%%%%%%%%%%%%%%%%%%%%%%%%%%%%%%%%%%%%
\bibliographystyle{plain}
\bibliography{references}

%%%%%%%%%%%%%%%%%%%%%%%%%%%%%%%%%%%%%%%%%%%%%%%
% Appendix
%%%%%%%%%%%%%%%%%%%%%%%%%%%%%%%%%%%%%%%%%%%%%%%
\newpage
\begin{appendices}
%%%%%%%%%%%%%%%%%%%%%%%%%%%%%%%%%%%%%%%%%%
\section{Denoising performance tables}
%%%%%%%%%%%%%%%%%%%%%%%%%%%%%%%%%%%%%%%%%%
\begin{table}[h!]
\centering
\small
\begin{tabular}{|c|c|c|c|c|c|c|c|}
\hline
{\bf Denoiser} & {\bf Mean} & {\bf Std Dev} & {\bf Min} & {\bf 25\%} & {\bf 50\%} & {\bf 75\%} & {\bf Max} \\
\hline
\hline
\begin{minipage}{17ex}\vspace{0.05in}Autoencoder\vspace{0.05in}\end{minipage} & 3114.94 & 25671.74 & 6.2 & 33.0 & 93.0 & 335.4 & 679435.1 \\
\hline\begin{minipage}{17ex}\vspace{0.05in}Butterworth\\ Filter\vspace{0.05in}\end{minipage} & 25723.42 & 91285.33 & 4.9 & 623.3 & 2238.4 & 10529.8 & 1657573.8 \\
\hline\begin{minipage}{17ex}\vspace{0.05in}Multi-Frequency Butterworth\vspace{0.05in}\end{minipage} & 25995.49 & 93009.65 & 4.3 & 602.3 & 2229.1 & 10637.6 & 1722930.9 \\
\hline\begin{minipage}{17ex}\vspace{0.05in}biosppy\vspace{0.05in}\end{minipage} & 24197.64 & 87116.53 & 2.2 & 439.9 & 1855.6 & 9821.3 & 1634449.4 \\
\hline\begin{minipage}{17ex}\vspace{0.05in}elgendi2010\vspace{0.05in}\end{minipage} & 27984.69 & 95833.63 & 23.2 & 1286.3 & 3392.7 & 12362.1 & 1716402.2 \\
\hline\begin{minipage}{17ex}\vspace{0.05in}engzeemod2012\vspace{0.05in}\end{minipage} & {\bf 1133.45} & {\bf11067.91} & {\bf 0.6} & {\bf 26.6} & {\bf 67.7} & {\bf 180.4} & {\bf 343080.1} \\
\hline\begin{minipage}{17ex}\vspace{0.05in}hamilton2002\vspace{0.05in}\end{minipage} & 25829.91 & 89215.30 & 19.4 & 1001.9 & 2770.5 & 10966.3 & 1579078.6 \\
\hline\begin{minipage}{17ex}\vspace{0.05in}neurokit\vspace{0.05in}\end{minipage} & 24421.97 & 87590.32 & 9.2 & 589.0 & 2118.4 & 10242.7 & 1459737.0 \\
\hline\begin{minipage}{17ex}\vspace{0.05in}pantompkins1985\vspace{0.05in}\end{minipage} & 25174.61 & 87329.14 & 18.8 & 918.2 & 2579.5 & 10490.5 & 1543653.1 \\
\hline\begin{minipage}{17ex}\vspace{0.05in}vg\vspace{0.05in}\end{minipage} & 24808.76 & 88917.35 & 3.3 & 530.5 & 2069.3 & 10023.1 & 1749421.9 \\
\hline
\end{tabular}
\caption{Statistical summary of the sum of squared differences ($SSD$) measured between denoised outputs and clean reference ECG signals for clean input data, reported for each denoising method.}
\label{table:cleaned_SSD}
\end{table}

\begin{table}[h!]
\centering\small
\begin{tabular}{|c|c|c|c|c|c|c|c|}
\hline
{\bf Denoiser} & {\bf Mean} & {\bf Std Dev} & {\bf Min} & {\bf 25\%} & {\bf 50\%} & {\bf 75\%} & {\bf Max} \\
\hline
\hline
\begin{minipage}{17ex}\vspace{0.05in}Autoencoder\vspace{0.05in}\end{minipage} & 1.97 & 2.22 & 0.1 & 0.6 & 1.3 & 2.4 & 16.0 \\
\hline\begin{minipage}{17ex}\vspace{0.05in}Butterworth\\ Filter\vspace{0.05in}\end{minipage} & 2.35 & 2.75 & 0.1 & 0.7 & 1.3 & 2.7 & 17.4 \\
\hline\begin{minipage}{17ex}\vspace{0.05in}Multi-Frequency Butterworth\vspace{0.05in}\end{minipage} & 2.34 & 2.70 & 0.1 & 0.7 & 1.3 & 2.7 & 17.7 \\
\hline\begin{minipage}{17ex}\vspace{0.05in}biosppy\vspace{0.05in}\end{minipage} & 2.29 & 2.73 & 0.1 & 0.7 & 1.3 & 2.6 & 21.4 \\
\hline\begin{minipage}{17ex}\vspace{0.05in}elgendi2010\vspace{0.05in}\end{minipage} & 3.63 & 2.89 & 0.3 & 2.0 & 2.7 & 4.1 & 21.8 \\
\hline\begin{minipage}{17ex}\vspace{0.05in}engzeemod2012\vspace{0.05in}\end{minipage} & {\bf 0.75} & {\bf 1.20} & {\bf 0.0} & {\bf 0.3} & {\bf 0.5} & {\bf 0.8} & {\bf 13.4} \\
\hline\begin{minipage}{17ex}\vspace{0.05in}hamilton2002\vspace{0.05in}\end{minipage} & 3.26 & 2.69 & 0.2 & 1.7 & 2.4 & 3.7 & 20.1 \\
\hline\begin{minipage}{17ex}\vspace{0.05in}neurokit\vspace{0.05in}\end{minipage} & 2.77 & 2.72 & 0.2 & 1.2 & 1.8 & 3.2 & 23.7 \\
\hline\begin{minipage}{17ex}\vspace{0.05in}pantompkins1985\vspace{0.05in}\end{minipage} & 3.16 & 2.69 & 0.2 & 1.6 & 2.3 & 3.5 & 20.5 \\
\hline\begin{minipage}{17ex}\vspace{0.05in}vg\vspace{0.05in}\end{minipage} & 2.24 & 2.64 & {\bf 0.0} & 0.7 & 1.2 & 2.5 & 16.0 \\
\hline
\end{tabular}
\caption{Statistical summary of the maximum absolute difference ($MAD$) measured between denoised outputs and clean reference ECG signals for clean input data, reported for each denoising method.}
\label{table:cleaned_MAD}
\end{table}

\begin{table}[h!]
\centering
\small
\begin{tabular}{|c|c|c|c|c|c|c|c|}
\hline
{\bf Denoiser} & {\bf Mean} & {\bf Std Dev} & {\bf Min} & {\bf 25\%} & {\bf 50\%} & {\bf 75\%} & {\bf Max} \\
\hline
\hline
\begin{minipage}{17ex}\vspace{0.05in}Autoencoder\vspace{0.05in}\end{minipage} & {\bf 0.01} & {\bf 0.02} & {\bf 0.001} & 0.006 & 0.008 & {\bf 0.012} & {\bf 0.423} \\
\hline\begin{minipage}{17ex}\vspace{0.05in}Butterworth\\ Filter\vspace{0.05in}\end{minipage} & 0.29 & 0.18 & 0.005 & 0.161 & 0.252 & 0.368 & 1.52 \\
\hline\begin{minipage}{17ex}\vspace{0.05in}Multi-Frequency Butterworth\vspace{0.05in}\end{minipage} & 0.29 & 0.19 & 0.015 & 0.155 & 0.248 & 0.37 & 1.324 \\
\hline\begin{minipage}{17ex}\vspace{0.05in}biosppy\vspace{0.05in}\end{minipage} & 0.25 & 0.16 & 0.003 & 0.133 & 0.224 & 0.333 & 0.809 \\
\hline\begin{minipage}{17ex}\vspace{0.05in}elgendi2010\vspace{0.05in}\end{minipage} & 0.44 & 0.16 & 0.046 & 0.325 & 0.402 & 0.528 & 1.051 \\
\hline\begin{minipage}{17ex}\vspace{0.05in}engzeemod2012\vspace{0.05in}\end{minipage} & 0.02 & 0.04 & {\bf 0.001} & {\bf 0.002} & {\bf 0.005} & 0.017 & 0.605 \\
\hline\begin{minipage}{17ex}\vspace{0.05in}hamilton2002\vspace{0.05in}\end{minipage} & 0.35 & 0.13 & 0.044 & 0.252 & 0.314 & 0.418 & 0.862 \\
\hline\begin{minipage}{17ex}\vspace{0.05in}neurokit\vspace{0.05in}\end{minipage} & 0.29 & 0.20 & 0.006 & 0.158 & 0.246 & 0.367 & 1.496 \\
\hline\begin{minipage}{17ex}\vspace{0.05in}pantompkins1985\vspace{0.05in}\end{minipage} & 0.33 & 0.13 & 0.043 & 0.23 & 0.293 & 0.394 & 0.842 \\
\hline\begin{minipage}{17ex}\vspace{0.05in}vg\vspace{0.05in}\end{minipage} & 0.28 & 0.20 & 0.004 & 0.135 & 0.233 & 0.351 & 1.423 \\
\hline
\end{tabular}
\caption{Statistical summary of the cosine distance measured between denoised outputs and clean reference ECG signals for clean input data, reported for each denoising method.}
\label{table:cleaned_COSDIST}
\end{table}

%%%%%%%%%%%%%%%%%%%%%%%%%%%%%%%%%%%%%%%%%%
%\section{Denoised noisy signals: denoising performance tables}
%%%%%%%%%%%%%%%%%%%%%%%%%%%%%%%%%%%%%%%%%%
%%%%%%%%%%%%%%%%%%%%%%%%%%%%%%%%%%%%%%
% LinearWander
%%%%%%%%%%%%%%%%%%%%%%%%%%%%%%%%%%%%%%
\begin{table}[h!]
\centering
\small
\begin{tabular}{|c|c|c|c|c|c|c|c|}
\hline
{\bf Denoiser} & {\bf Mean} & {\bf Std Dev} & {\bf Min} & {\bf 25\%} & {\bf 50\%} & {\bf 75\%} & {\bf Max} \\
\hline
\hline
\begin{minipage}{17ex}\vspace{0.05in}Autoencoder\vspace{0.05in}\end{minipage} & {\bf 9917.30} & {\bf 32886.79} & 100.4 & 731.1 & 2150.4 & {\bf 7222.7} & 787130.1 \\
\hline\begin{minipage}{17ex}\vspace{0.05in}Butterworth\\ Filter\vspace{0.05in}\end{minipage} & 25836.81 & 91318.57 & 6.5 & 731.8 & 2345.5 & 10618.9 & 1658112.6 \\
\hline\begin{minipage}{17ex}\vspace{0.05in}Multi-Frequency Butterworth\vspace{0.05in}\end{minipage} & 26096.33 & 93038.41 & 5.7 & 695.0 & 2289.3 & 10800.2 & 1723393.2 \\
\hline\begin{minipage}{17ex}\vspace{0.05in}biosppy\vspace{0.05in}\end{minipage} & 24837.91 & 87291.12 & 10.2 & 576.8 & 2819.9 & 10597.0 & 1622792.4 \\
\hline\begin{minipage}{17ex}\vspace{0.05in}elgendi2010\vspace{0.05in}\end{minipage} & 28057.22 & 95852.60 & 23.7 & 1338.6 & 3453.6 & 12416.7 & 1714080.1 \\
\hline\begin{minipage}{17ex}\vspace{0.05in}engzeemod2012\vspace{0.05in}\end{minipage} & 35150.48 & 68120.76 & 91.0 & {\bf 553.1} & {\bf 1771.5} & 18598.6 & {\bf 513433.2} \\
\hline\begin{minipage}{17ex}\vspace{0.05in}hamilton2002\vspace{0.05in}\end{minipage} & 26046.89 & 89294.19 & 19.8 & 1078.2 & 2886.3 & 11162.2 & 1554455.0 \\
\hline\begin{minipage}{17ex}\vspace{0.05in}neurokit\vspace{0.05in}\end{minipage} & 25469.66 & 87205.08 & 22.7 & 759.4 & 3612.0 & 12740.4 & 1478642.4 \\
\hline\begin{minipage}{17ex}\vspace{0.05in}pantompkins1985\vspace{0.05in}\end{minipage} & 25503.45 & 87418.26 & 19.8 & 1060.9 & 2810.4 & 10893.0 & 1514134.5 \\
\hline\begin{minipage}{17ex}\vspace{0.05in}vg\vspace{0.05in}\end{minipage} & 24937.71 & 88954.85 & {\bf 5.1} & 671.5 & 2188.5 & 10089.7 & 1750013.4 \\
\hline
\end{tabular}
\caption{Statistical summary of the sum of squared differences ($SSD$) measured between denoised outputs and clean reference ECG signals for noisy input data corrupted with LinearWander noise, reported for each denoising method.}
\label{table:LinearWandercleaned_SSD}
\end{table}

\begin{table}[h!]
\centering
\small
\begin{tabular}{|c|c|c|c|c|c|c|c|}
\hline
{\bf Denoiser} & {\bf Mean} & {\bf Std Dev} & {\bf Min} & {\bf 25\%} & {\bf 50\%} & {\bf 75\%} & {\bf Max} \\
\hline
\hline
\begin{minipage}{17ex}\vspace{0.05in}Autoencoder\vspace{0.05in}\end{minipage} & {\bf 2.62} & {\bf 2.26} & 0.3 & 1.1 & 1.9 & 3.6 & {\bf 17.0} \\
\hline\begin{minipage}{17ex}\vspace{0.05in}Butterworth\\ Filter\vspace{0.05in}\end{minipage} & 2.91 & 2.89 & {\bf 0.2} & {\bf 0.9} & 1.8 & 4.0 & 19.4 \\
\hline\begin{minipage}{17ex}\vspace{0.05in}Multi-Frequency Butterworth\vspace{0.05in}\end{minipage} & 2.99 & 2.92 & {\bf 0.2} & {\bf 0.9} & 1.8 & 4.3 & 19.5 \\
\hline\begin{minipage}{17ex}\vspace{0.05in}biosppy\vspace{0.05in}\end{minipage} & 2.97 & 2.99 & 0.3 & {\bf 0.9} & 1.7 & 4.4 & 21.4 \\
\hline\begin{minipage}{17ex}\vspace{0.05in}elgendi2010\vspace{0.05in}\end{minipage} & 3.75 & 2.89 & 0.3 & 2.1 & 2.9 & 4.3 & 21.8 \\
\hline\begin{minipage}{17ex}\vspace{0.05in}engzeemod2012\vspace{0.05in}\end{minipage} & 2.99 & 3.23 & 0.3 & 1.0 & {\bf 1.4} & {\bf 3.4} & 19.5 \\
\hline\begin{minipage}{17ex}\vspace{0.05in}hamilton2002\vspace{0.05in}\end{minipage} & 3.48 & 2.69 & {\bf 0.2} & 1.8 & 2.7 & 4.1 & 20.1 \\
\hline\begin{minipage}{17ex}\vspace{0.05in}neurokit\vspace{0.05in}\end{minipage} & 3.28 & 2.87 & 0.3 & 1.4 & 2.2 & 4.5 & 24.3 \\
\hline\begin{minipage}{17ex}\vspace{0.05in}pantompkins1985\vspace{0.05in}\end{minipage} & 3.56 & 2.77 & 0.3 & 1.7 & 2.6 & 4.7 & 20.7 \\
\hline\begin{minipage}{17ex}\vspace{0.05in}vg\vspace{0.05in}\end{minipage} & 2.82 & 2.82 & {\bf 0.2} & {\bf 0.9} & 1.7 & 4.0 & 18.7 \\
\hline
\end{tabular}
\caption{Statistical summary of the maximum absolute difference ($MAD$) measured between denoised outputs and clean reference ECG signals for noisy input data corrupted with LinearWander noise, reported for each denoising method.}
\label{table:LinearWandercleaned_MAD}
\end{table}

\begin{table}[h!]
\centering
\small
\begin{tabular}{|c|c|c|c|c|c|c|c|}
\hline
{\bf Denoiser} & {\bf Mean} & {\bf Std Dev} & {\bf Min} & {\bf 25\%} & {\bf 50\%} & {\bf 75\%} & {\bf Max} \\
\hline
\hline
\begin{minipage}{17ex}\vspace{0.05in}Autoencoder\vspace{0.05in}\end{minipage} & {\bf 0.24} & 0.25 & 0.002 & 0.048 & {\bf 0.146} & {\bf 0.346} & 1.398 \\
\hline\begin{minipage}{17ex}\vspace{0.05in}Butterworth\\ Filter\vspace{0.05in}\end{minipage} & 0.31 & 0.18 & 0.006 & 0.175 & 0.266 & 0.39 & 1.515 \\
\hline\begin{minipage}{17ex}\vspace{0.05in}Multi-Frequency Butterworth\vspace{0.05in}\end{minipage} & 0.30 & 0.19 & 0.015 & 0.169 & 0.264 & 0.389 & 1.32 \\
\hline\begin{minipage}{17ex}\vspace{0.05in}biosppy\vspace{0.05in}\end{minipage} & 0.30 & 0.19 & 0.004 & 0.155 & 0.259 & 0.412 & {\bf 0.982} \\
\hline\begin{minipage}{17ex}\vspace{0.05in}elgendi2010\vspace{0.05in}\end{minipage} & 0.45 & 0.16 & 0.048 & 0.328 & 0.408 & 0.543 & 1.042 \\
\hline\begin{minipage}{17ex}\vspace{0.05in}engzeemod2012\vspace{0.05in}\end{minipage} & 0.28 & 0.31 & {\bf 0.000} & {\bf 0.047} & 0.155 & 0.417 & 1.53 \\
\hline\begin{minipage}{17ex}\vspace{0.05in}hamilton2002\vspace{0.05in}\end{minipage} & 0.37 & {\bf 0.15} & 0.045 & 0.257 & 0.322 & 0.45 & 1.212 \\
\hline\begin{minipage}{17ex}\vspace{0.05in}neurokit\vspace{0.05in}\end{minipage} & 0.35 & 0.24 & 0.006 & 0.181 & 0.284 & 0.47 & 1.493 \\
\hline\begin{minipage}{17ex}\vspace{0.05in}pantompkins1985\vspace{0.05in}\end{minipage} & 0.36 & 0.16 & 0.044 & 0.238 & 0.307 & 0.442 & 1.207 \\
\hline\begin{minipage}{17ex}\vspace{0.05in}vg\vspace{0.05in}\end{minipage} & 0.29 & 0.20 & 0.005 & 0.148 & 0.252 & 0.379 & 1.422 \\
\hline
\end{tabular}
\caption{Statistical summary of the cosine distance measured between denoised outputs and clean reference ECG signals for noisy input data corrupted with LinearWander noise, reported for each denoising method.}
\label{table:LinearWandercleaned_COSDIST}
\end{table}

%%%%%%%%%%%%%%%%%%%%%%%%%%%%%%%%%%%%%%
% SineWander
%%%%%%%%%%%%%%%%%%%%%%%%%%%%%%%%%%%%%%
\begin{table}[h!]
\centering
\small
\begin{tabular}{|c|c|c|c|c|c|c|c|}
\hline
{\bf Denoiser} & {\bf Mean} & {\bf Std Dev} & {\bf Min} & {\bf 25\%} & {\bf 50\%} & {\bf 75\%} & {\bf Max} \\
\hline
\hline
\begin{minipage}{17ex}\vspace{0.05in}Autoencoder\vspace{0.05in}\end{minipage} & 8282.37 & 25652.69 & 5069.5 & 5152.7 & {\bf 5221.7} & {\bf 5520.4} & 687045.4 \\
\hline\begin{minipage}{17ex}\vspace{0.05in}Butterworth\\ Filter\vspace{0.05in}\end{minipage} & 30244.65 & 91285.32 & 4526.1 & 5144.5 & 6759.7 & 15050.6 & 1662095.0 \\
\hline\begin{minipage}{17ex}\vspace{0.05in}Multi-Frequency Butterworth\vspace{0.05in}\end{minipage} & 30516.66 & 93009.64 & 4525.5 & 5123.5 & 6750.3 & 15158.9 & 1727451.9 \\
\hline\begin{minipage}{17ex}\vspace{0.05in}biosppy\vspace{0.05in}\end{minipage} & 28730.56 & 87115.86 & {\bf 4535.0} & {\bf 4971.3} & 6387.5 & 14326.7 & 1638947.9 \\
\hline\begin{minipage}{17ex}\vspace{0.05in}elgendi2010\vspace{0.05in}\end{minipage} & 32505.62 & 95833.39 & 4544.1 & 5807.3 & 7914.2 & 16880.7 & 1720892.0 \\
\hline\begin{minipage}{17ex}\vspace{0.05in}engzeemod2012\vspace{0.05in}\end{minipage} & {\bf 6630.07} & {\bf 11068.27} & 5487.8 & 5523.1 & 5564.6 & 5676.5 & {\bf 348507.9} \\
\hline\begin{minipage}{17ex}\vspace{0.05in}hamilton2002\vspace{0.05in}\end{minipage} & 30351.46 & 89213.19 & 4541.2 & 5523.3 & 7288.9 & 15500.5 & 1583304.5 \\
\hline\begin{minipage}{17ex}\vspace{0.05in}neurokit\vspace{0.05in}\end{minipage} & 28979.16 & 87664.78 & 4554.7 & 5146.6 & 6647.3 & 14701.2 & 1469157.6 \\
\hline\begin{minipage}{17ex}\vspace{0.05in}pantompkins1985\vspace{0.05in}\end{minipage} & 29698.63 & 87326.91 & 4542.9 & 5439.3 & 7102.4 & 15024.9 & 1547815.5 \\
\hline\begin{minipage}{17ex}\vspace{0.05in}vg\vspace{0.05in}\end{minipage} & 29330.08 & 88917.35 & {\bf 4524.7} & 5051.8 & 6590.9 & 14544.5 & 1753943.1 \\
\hline
\end{tabular}
\caption{Statistical summary of the sum of squared differences ($SSD$) measured between denoised outputs and clean reference ECG signals for noisy input data corrupted with SineWander noise, reported for each denoising method.}
\label{table:SineWandercleaned_SSD}
\end{table}

\begin{table}[h!]
\centering
\small
\begin{tabular}{|c|c|c|c|c|c|c|c|}
\hline
{\bf Denoiser} & {\bf Mean} & {\bf Std Dev} & {\bf Min} & {\bf 25\%} & {\bf 50\%} & {\bf 75\%} & {\bf Max} \\
\hline
\hline
\begin{minipage}{17ex}\vspace{0.05in}Autoencoder\vspace{0.05in}\end{minipage} & 2.56 & 1.91 & {\bf 1.8} & {\bf 1.8} & {\bf 1.8} & 2.4 & 16.0 \\
\hline\begin{minipage}{17ex}\vspace{0.05in}Butterworth\\ Filter\vspace{0.05in}\end{minipage} & 2.88 & 2.47 & {\bf 1.8} & {\bf 1.8} & {\bf 1.8} & 2.7 & 17.4 \\
\hline\begin{minipage}{17ex}\vspace{0.05in}Multi-Frequency Butterworth\vspace{0.05in}\end{minipage} & 2.87 & 2.42 & {\bf 1.8} & {\bf 1.8} & {\bf 1.8} & 2.7 & 17.7 \\
\hline\begin{minipage}{17ex}\vspace{0.05in}biosppy\vspace{0.05in}\end{minipage} & 2.83 & 2.46 & {\bf 1.8} & {\bf 1.8} & {\bf 1.8} & 2.6 & 21.4 \\
\hline\begin{minipage}{17ex}\vspace{0.05in}elgendi2010\vspace{0.05in}\end{minipage} & 3.72 & 2.82 & {\bf 1.8} & 2.0 & 2.7 & 4.1 & 21.8 \\
\hline\begin{minipage}{17ex}\vspace{0.05in}engzeemod2012\vspace{0.05in}\end{minipage} & {\bf 1.91} & {\bf 0.99} & {\bf 1.8} & {\bf 1.8} & {\bf 1.8} & {\bf 1.8} & {\bf 13.8} \\
\hline\begin{minipage}{17ex}\vspace{0.05in}hamilton2002\vspace{0.05in}\end{minipage} & 3.39 & 2.60 & {\bf 1.8} & {\bf 1.8} & 2.4 & 3.7 & 20.0 \\
\hline\begin{minipage}{17ex}\vspace{0.05in}neurokit\vspace{0.05in}\end{minipage} & 3.05 & 2.56 & {\bf 1.8} & {\bf 1.8} & {\bf 1.8} & 3.2 & 23.7 \\
\hline\begin{minipage}{17ex}\vspace{0.05in}pantompkins1985\vspace{0.05in}\end{minipage} & 3.30 & 2.59 & {\bf 1.8} & {\bf 1.8} & 2.3 & 3.5 & 20.5 \\
\hline\begin{minipage}{17ex}\vspace{0.05in}vg\vspace{0.05in}\end{minipage} & 2.81 & 2.35 & {\bf 1.8} & {\bf 1.8} & {\bf 1.8} & 2.6 & 16.0 \\
\hline
\end{tabular}
\caption{Statistical summary of the maximum absolute difference ($MAD$) measured between denoised outputs and clean reference ECG signals for noisy input data corrupted with SineWander noise, reported for each denoising method.}
\label{table:SineWandercleaned_MAD}
\end{table}

\begin{table}[h!]
\centering
\small
\begin{tabular}{|c|c|c|c|c|c|c|c|}
\hline
{\bf Denoiser} & {\bf Mean} & {\bf Std Dev} & {\bf Min} & {\bf 25\%} & {\bf 50\%} & {\bf 75\%} & {\bf Max} \\
\hline
\hline
\begin{minipage}{17ex}\vspace{0.05in}Autoencoder\vspace{0.05in}\end{minipage} & {\bf 0.32} & 0.21 & 0.004 & {\bf 0.132} & {\bf 0.311} & {\bf 0.491} & 0.953 \\
\hline\begin{minipage}{17ex}\vspace{0.05in}Butterworth\\ Filter\vspace{0.05in}\end{minipage} & 0.57 & {\bf 0.18} & 0.013 & 0.431 & 0.574 & 0.706 & 1.286 \\
\hline\begin{minipage}{17ex}\vspace{0.05in}Multi-Frequency Butterworth\vspace{0.05in}\end{minipage} & 0.57 & {\bf 0.18} & 0.04 & 0.433 & 0.576 & 0.708 & 1.152 \\
\hline\begin{minipage}{17ex}\vspace{0.05in}biosppy\vspace{0.05in}\end{minipage} & 0.54 & {\bf 0.18} & 0.044 & 0.397 & 0.55 & 0.685 & 0.975 \\
\hline\begin{minipage}{17ex}\vspace{0.05in}elgendi2010\vspace{0.05in}\end{minipage} & 0.67 & 0.19 & 0.051 & 0.531 & 0.705 & 0.822 & 1.045 \\
\hline\begin{minipage}{17ex}\vspace{0.05in}engzeemod2012\vspace{0.05in}\end{minipage} & 0.33 & 0.22 & {\bf 0.002} & 0.136 & 0.327 & 0.509 & {\bf 0.952} \\
\hline\begin{minipage}{17ex}\vspace{0.05in}hamilton2002\vspace{0.05in}\end{minipage} & 0.63 & {\bf 0.18} & 0.048 & 0.487 & 0.655 & 0.773 & 1.015 \\
\hline\begin{minipage}{17ex}\vspace{0.05in}neurokit\vspace{0.05in}\end{minipage} & 0.57 & 0.20 & 0.043 & 0.415 & 0.581 & 0.72 & 1.46 \\
\hline\begin{minipage}{17ex}\vspace{0.05in}pantompkins1985\vspace{0.05in}\end{minipage} & 0.61 & {\bf 0.18} & 0.047 & 0.467 & 0.631 & 0.754 & 1.003 \\
\hline\begin{minipage}{17ex}\vspace{0.05in}vg\vspace{0.05in}\end{minipage} & 0.56 & 0.19 & 0.012 & 0.412 & 0.562 & 0.701 & 1.348 \\
\hline
\end{tabular}
\caption{Statistical summary of the cosine distance measured between denoised outputs and clean reference ECG signals for noisy input data corrupted with SineWander noise, reported for each denoising method.}
\label{table:SineWandercleaned_COSDIST}
\end{table}

%%%%%%%%%%%%%%%%%%%%%%%%%%%%%%%%%%%%%%
% Multi-noise
%%%%%%%%%%%%%%%%%%%%%%%%%%%%%%%%%%%%%%
\begin{table}[h!]
\centering
\footnotesize
\begin{tabular}{|c|c|c|c|c|c|c|c|}
\hline
{\bf Denoiser} & {\bf Mean} & {\bf Std Dev} & {\bf Min} & {\bf 25\%} & {\bf 50\%} & {\bf 75\%} & {\bf Max} \\
\hline
\hline
\begin{minipage}{17ex}\vspace{0.05in}Autoencoder\vspace{0.05in}\end{minipage} & 47361.93 & 68297.40 & 12010.7 & 14004.9 & 16105.7 & 32168.7 & 860817.2 \\
\hline\begin{minipage}{17ex}\vspace{0.05in}Butterworth\\ Filter\vspace{0.05in}\end{minipage} & 69970.40 & 114604.94 & 12055.2 & 15636.3 & 21094.5 & 81272.6 & 1672085.5 \\
\hline\begin{minipage}{17ex}\vspace{0.05in}Multi-Frequency Butterworth\vspace{0.05in}\end{minipage} & 70242.47 & 116029.32 & 12053.9 & 15613.5 & 21110.0 & 81979.3 & 1737442.5 \\
\hline\begin{minipage}{17ex}\vspace{0.05in}biosppy\vspace{0.05in}\end{minipage} & 68444.62 & 110806.56 & 12051.8 & 15398.3 & 20567.2 & 79357.7 & 1648961.0 \\
\hline\begin{minipage}{17ex}\vspace{0.05in}elgendi2010\vspace{0.05in}\end{minipage} & 72231.67 & 118557.33 & 12329.0 & 16441.0 & 22458.3 & 84970.7 & 1730914.0 \\
\hline\begin{minipage}{17ex}\vspace{0.05in}engzeemod2012\vspace{0.05in}\end{minipage} & {\bf 45380.43} & {\bf 62543.32} & {\bf 12010.4} & {\bf 13918.0} & {\bf 16035.6} & {\bf 30532.6} & {\bf 641887.4} \\
\hline\begin{minipage}{17ex}\vspace{0.05in}hamilton2002\vspace{0.05in}\end{minipage} & 70076.90 & 112924.53 & 12211.7 & 16031.9 & 21608.7 & 82106.0 & 1593590.2 \\
\hline\begin{minipage}{17ex}\vspace{0.05in}neurokit\vspace{0.05in}\end{minipage} & 68668.95 & 110513.15 & 12111.9 & 15548.8 & 20839.3 & 81857.1 & 1526297.9 \\
\hline\begin{minipage}{17ex}\vspace{0.05in}{pantompkins1985}\vspace{0.05in}\end{minipage} & 69421.60 & 111320.97 & 12179.6 & 15917.4 & 21340.5 & 80903.5 & 1558164.8 \\
\hline\begin{minipage}{17ex}\vspace{0.05in}vg\vspace{0.05in}\end{minipage} & 69055.74 & 112389.58 & 12047.9 & 15491.8 & 20766.7 & 79320.8 & 1763933.5 \\
\hline
\end{tabular}
\caption{Statistical summary of the sum of squared differences ($SSD$) measured between denoised outputs and clean reference ECG signals for noisy input data corrupted with LinearWander, MuscleArtifact, ShockPulses, and SineWander noise, reported for each denoising method.}
\label{table:multinoisecleaned_SSD}
\end{table}

\begin{table}[h!]
\centering
\small
\begin{tabular}{|c|c|c|c|c|c|c|c|}
\hline
{\bf Denoiser} & {\bf Mean} & {\bf Std Dev} & {\bf Min} & {\bf 25\%} & {\bf 50\%} & {\bf 75\%} & {\bf Max} \\
\hline
\hline
\begin{minipage}{17ex}\vspace{0.05in}Autoencoder\vspace{0.05in}\end{minipage} & {\bf 17.11} & {\bf 2.53} & {\bf 15.0} & {\bf 15.6} & {\bf 15.9} & {\bf 17.2} & {\bf 29.6} \\
\hline\begin{minipage}{17ex}\vspace{0.05in}Butterworth\\ Filter\vspace{0.05in}\end{minipage} & {\bf 17.11} & {\bf 2.53} & {\bf 15.0} & {\bf 15.6} & {\bf 15.9} & {\bf 17.2} & {\bf 29.6} \\
\hline\begin{minipage}{17ex}\vspace{0.05in}Multi-Frequency Butterworth\vspace{0.05in}\end{minipage} & {\bf 17.11} & {\bf 2.53} & {\bf 15.0} & {\bf 15.6} & {\bf 15.9} & {\bf 17.2} & {\bf 29.6} \\
\hline\begin{minipage}{17ex}\vspace{0.05in}biosppy\vspace{0.05in}\end{minipage} & 17.12 & {\bf 2.53} & {\bf 15.0} & {\bf 15.6} & {\bf 15.9} & {\bf 17.2} & {\bf 29.6} \\
\hline\begin{minipage}{17ex}\vspace{0.05in}elgendi2010\vspace{0.05in}\end{minipage} & 17.12 & {\bf 2.53} & {\bf 15.0} & {\bf 15.6} & {\bf 15.9} & {\bf 17.2} & {\bf 29.6} \\
\hline\begin{minipage}{17ex}\vspace{0.05in}engzeemod2012\vspace{0.05in}\end{minipage} & {\bf 17.11} & {\bf 2.53} & {\bf 15.0} & {\bf 15.6} & {\bf 15.9} & {\bf 17.2} & {\bf 29.6} \\
\hline\begin{minipage}{17ex}\vspace{0.05in}hamilton2002\vspace{0.05in}\end{minipage} & {\bf 17.11} & {\bf 2.53} & {\bf 15.0} & {\bf 15.6} & {\bf 15.9} & {\bf 17.2} & {\bf 29.6} \\
\hline\begin{minipage}{17ex}\vspace{0.05in}neurokit\vspace{0.05in}\end{minipage} & 17.12 & {\bf 2.53} & {\bf 15.0} & {\bf 15.6} & {\bf 15.9} & {\bf 17.2} & {\bf 29.6} \\
\hline\begin{minipage}{17ex}\vspace{0.05in}pantompkins1985\vspace{0.05in}\end{minipage} & 17.12 & {\bf 2.53} & {\bf 15.0} & {\bf 15.6} & {\bf 15.9} & {\bf 17.2} & {\bf 29.6} \\
\hline\begin{minipage}{17ex}\vspace{0.05in}vg\vspace{0.05in}\end{minipage} & {\bf 17.11} & {\bf 2.53} & {\bf 15.0} & {\bf 15.6} & {\bf 15.9} & {\bf 17.2} & {\bf 29.6} \\
\hline
\end{tabular}
\caption{Statistical summary of the maximum absolute difference ($MAD$) measured between denoised outputs and clean reference ECG signals for noisy input data corrupted with LinearWander, MuscleArtifact, ShockPulses, and SineWander noise, reported for each denoising method.}
\label{table:multinoisecleaned_MAD}
\end{table}

\begin{table}[h!]
\centering
\small
\begin{tabular}{|c|c|c|c|c|c|c|c|}
\hline
{\bf Denoiser} & {\bf Mean} & {\bf Std Dev} & {\bf Min} & {\bf 25\%} & {\bf 50\%} & {\bf 75\%} & {\bf Max} \\
\hline
\hline
\begin{minipage}{17ex}\vspace{0.05in}Autoencoder\vspace{0.05in}\end{minipage} & 0.55 & 0.29 & 0.008 & 0.325 & {\bf 0.551} & {\bf 0.745} & 1.58 \\
\hline\begin{minipage}{17ex}\vspace{0.05in}Butterworth\\ Filter\vspace{0.05in}\end{minipage} & 0.75 & {\bf 0.24} & 0.103 & 0.581 & 0.746 & 0.9 & 1.63 \\
\hline\begin{minipage}{17ex}\vspace{0.05in}Multi-Frequency Butterworth\vspace{0.05in}\end{minipage} & 0.75 & {\bf 0.24} & 0.093 & 0.583 & 0.747 & 0.902 & 1.63 \\
\hline\begin{minipage}{17ex}\vspace{0.05in}biosppy\vspace{0.05in}\end{minipage} & 0.73 & {\bf 0.24} & 0.094 & 0.555 & 0.726 & 0.88 & 1.626 \\
\hline\begin{minipage}{17ex}\vspace{0.05in}elgendi2010\vspace{0.05in}\end{minipage} & 0.81 & {\bf 0.24} & 0.104 & 0.655 & 0.824 & 0.956 & 1.64 \\
\hline\begin{minipage}{17ex}\vspace{0.05in}engzeemod2012\vspace{0.05in}\end{minipage} & {\bf 0.54} & 0.30 & {\bf 0.005} & {\bf 0.32} & 0.554 & 0.746 & {\bf 1.576} \\
\hline\begin{minipage}{17ex}\vspace{0.05in}hamilton2002\vspace{0.05in}\end{minipage} & 0.78 & {\bf 0.24} & 0.099 & 0.625 & 0.793 & 0.931 & 1.636 \\
\hline\begin{minipage}{17ex}\vspace{0.05in}neurokit\vspace{0.05in}\end{minipage} & 0.74 & 0.25 & 0.071 & 0.562 & 0.742 & 0.898 & 1.634 \\
\hline\begin{minipage}{17ex}\vspace{0.05in}pantompkins1985\vspace{0.05in}\end{minipage} & 0.77 & {\bf 0.24} & 0.098 & 0.608 & 0.778 & 0.92 & 1.634 \\
\hline\begin{minipage}{17ex}\vspace{0.05in}vg\vspace{0.05in}\end{minipage} & 0.74 & 0.25 & 0.08 & 0.564 & 0.736 & 0.896 & 1.629 \\
\hline
\end{tabular}
\caption{Statistical summary of the cosine distance measured between denoised outputs and clean reference ECG signals for noisy input data corrupted with LinearWander, MuscleArtifact, ShockPulses, and SineWander noise, reported for each denoising method.}
\label{table:multinoisecleaned_COSDIST}
\end{table}

%%%%%%%%%%%%%%%%%%%%%%%%%%%%%%%%%%%%%%%%%%
\section{Delineation performance tables}
%%%%%%%%%%%%%%%%%%%%%%%%%%%%%%%%%%%%%%%%%%
\begin{table}[h!]
\centering
\small
\begin{tabular}{|c|c|c|c|c|c|c|c|}
\hline
{\bf Denoiser} & {\bf Mean} & {\bf Std Dev} & {\bf Min} & {\bf 25\%} & {\bf 50\%} & {\bf 75\%} & {\bf Max} \\
\hline
\hline
\begin{minipage}{17ex}\vspace{0.05in}Autoencoder\vspace{0.05in}\end{minipage} & {\bf 18.74} & {\bf 22.36} & {\bf 0} & {\bf 8} & {\bf 14} & {\bf 22} & {\bf 292} \\
\hline\begin{minipage}{17ex}\vspace{0.05in}Butterworth\\ Filter\vspace{0.05in}\end{minipage} & 37.12 & 50.49 & {\bf 0} & 10 & 20 & 41 & 633 \\
\hline\begin{minipage}{17ex}\vspace{0.05in}Multi-Frequency Butterworth\vspace{0.05in}\end{minipage} & 33.04 & 44.79 & {\bf 0} & 10 & 18 & 36 & 528 \\
\hline\begin{minipage}{17ex}\vspace{0.05in}biosppy\vspace{0.05in}\end{minipage} & 51.06 & 47.08 & {\bf 0} & 20 & 36 & 64 & 492 \\
\hline\begin{minipage}{17ex}\vspace{0.05in}elgendi2010\vspace{0.05in}\end{minipage} & 198.10 & 103.13 & {\bf 0} & 128 & 202 & 264 & 586 \\
\hline\begin{minipage}{17ex}\vspace{0.05in}engzeemod2012\vspace{0.05in}\end{minipage} & 64.98 & 46.84 & {\bf 0} & 29 & 56 & 90 & 354 \\
\hline\begin{minipage}{17ex}\vspace{0.05in}hamilton2002\vspace{0.05in}\end{minipage} & 172.42 & 93.25 & {\bf 0} & 108 & 171 & 230 & 596 \\
\hline\begin{minipage}{17ex}\vspace{0.05in}neurokit\vspace{0.05in}\end{minipage} & 90.48 & 65.09 & {\bf 0} & 47 & 74 & 117 & 536 \\
\hline\begin{minipage}{17ex}\vspace{0.05in}pantompkins1985\vspace{0.05in}\end{minipage} & 151.69 & 85.12 & {\bf 0} & 96 & 142 & 200 & 560 \\
\hline\begin{minipage}{17ex}\vspace{0.05in}vg\vspace{0.05in}\end{minipage} & 27.51 & 39.69 & {\bf 0} & {\bf 8} & {\bf 14} & 30 & 541 \\
\hline
\end{tabular}
\caption{Statistical summary of the additional ECG delineation error introduced by the denoising process for clean input signals, reported for each denoising method.}
\label{table:denoised_clean_delineation}
\end{table}

\begin{table}[h!]
\centering
\small
\begin{tabular}{|c|c|c|c|c|c|c|c|}
\hline
{\bf Denoiser} & {\bf Mean} & {\bf Std Dev} & {\bf Min} & {\bf 25\%} & {\bf 50\%} & {\bf 75\%} & {\bf Max} \\
\hline
\hline
\begin{minipage}{17ex}\vspace{0.05in}Autoencoder\vspace{0.05in}\end{minipage} & {\bf 33.41} & {\bf 33.19} & {\bf 0} & {\bf 14} & {\bf 24} & {\bf 40} & 426 \\
\hline\begin{minipage}{17ex}\vspace{0.05in}Butterworth\\ Filter\vspace{0.05in}\end{minipage} & 48.46 & 54.46 & {\bf 0} & 18 & 30 & 55 & 644 \\
\hline\begin{minipage}{17ex}\vspace{0.05in}Multi-Frequency Butterworth\vspace{0.05in}\end{minipage} & 43.49 & 48.53 & {\bf 0} & 16 & 28 & 48 & 506 \\
\hline\begin{minipage}{17ex}\vspace{0.05in}biosppy\vspace{0.05in}\end{minipage} & 48.67 & 39.80 & {\bf 0} & 24 & 38 & 60 & 498 \\
\hline\begin{minipage}{17ex}\vspace{0.05in}elgendi2010\vspace{0.05in}\end{minipage} & 239.35 & 85.68 & {\bf 0} & 180 & 234 & 291 & 593 \\
\hline\begin{minipage}{17ex}\vspace{0.05in}engzeemod2012\vspace{0.05in}\end{minipage} & 68.49 & 44.94 & {\bf 0} & 36 & 60 & 90 & {\bf 424} \\
\hline\begin{minipage}{17ex}\vspace{0.05in}hamilton2002\vspace{0.05in}\end{minipage} & 204.71 & 82.56 & {\bf 0} & 149 & 196 & 255 & 566 \\
\hline\begin{minipage}{17ex}\vspace{0.05in}neurokit\vspace{0.05in}\end{minipage} & 102.36 & 69.52 & {\bf 0} & 55 & 82 & 126 & 516 \\
\hline\begin{minipage}{17ex}\vspace{0.05in}pantompkins1985\vspace{0.05in}\end{minipage} & 179.05 & 79.26 & {\bf 0} & 124 & 166 & 222 & 556 \\
\hline\begin{minipage}{17ex}\vspace{0.05in}vg\vspace{0.05in}\end{minipage} & 37.23 & 43.12 & {\bf 0} & {\bf 14} & {\bf 24} & 42 & 487 \\
\hline
\end{tabular}
\caption{Statistical summary of the ECG delineation error measured after denoising for noisy input data corrupted with LinearWander noise, reported for each denoising method.}
\label{table:delineated_LinearWander}
\end{table}

\begin{table}[h!]
\centering
\small
\begin{tabular}{|c|c|c|c|c|c|c|c|}
\hline
{\bf Denoiser} & {\bf Mean} & {\bf Std Dev} & {\bf Min} & {\bf 25\%} & {\bf 50\%} & {\bf 75\%} & {\bf Max} \\
\hline
\hline
\begin{minipage}{17ex}\vspace{0.05in}Autoencoder\vspace{0.05in}\end{minipage} & {\bf 28.25} & {\bf 28.98} & {\bf 0} & 12 & {\bf 20} & {\bf 32} & {\bf 357} \\
\hline\begin{minipage}{17ex}\vspace{0.05in}Butterworth\\ Filter\vspace{0.05in}\end{minipage} & 44.22 & 53.67 & {\bf 0} & 14 & 26 & 50 & 615 \\
\hline\begin{minipage}{17ex}\vspace{0.05in}Multi-Frequency Butterworth\vspace{0.05in}\end{minipage} & 39.59 & 47.78 & {\bf 0} & 12 & 23 & 44 & 528 \\
\hline\begin{minipage}{17ex}\vspace{0.05in}biosppy\vspace{0.05in}\end{minipage} & 45.76 & 39.73 & {\bf 0} & 22 & 35 & 56 & 541 \\
\hline\begin{minipage}{17ex}\vspace{0.05in}elgendi2010\vspace{0.05in}\end{minipage} & 236.58 & 85.21 & {\bf 0} & 178 & 231 & 289 & 581 \\
\hline\begin{minipage}{17ex}\vspace{0.05in}engzeemod2012\vspace{0.05in}\end{minipage} & 64.17 & 42.40 & {\bf 0} & 32 & 56 & 86 & 346 \\
\hline\begin{minipage}{17ex}\vspace{0.05in}hamilton2002\vspace{0.05in}\end{minipage} & 201.11 & 82.24 & {\bf 0} & 145 & 192 & 249 & 592 \\
\hline\begin{minipage}{17ex}\vspace{0.05in}neurokit\vspace{0.05in}\end{minipage} & 98.07 & 70.20 & {\bf 0} & 50 & 76 & 122 & 508 \\
\hline\begin{minipage}{17ex}\vspace{0.05in}pantompkins1985\vspace{0.05in}\end{minipage} & 174.75 & 79.03 & {\bf 0} & 118 & 160 & 218 & 574 \\
\hline\begin{minipage}{17ex}\vspace{0.05in}vg\vspace{0.05in}\end{minipage} & 33.95 & 42.73 & {\bf 0} & {\bf 10} & {\bf 20} & 38 & 543 \\
\hline
\end{tabular}
\caption{Statistical summary of the ECG delineation error measured after denoising for noisy input data corrupted with SineWander noise, reported for each denoising method.}
\label{table:delineated_SineWander}
\end{table}

\begin{table}[h!]
\centering
\small
\begin{tabular}{|c|c|c|c|c|c|c|c|}
\hline
{\bf Denoiser} & {\bf Mean} & {\bf Std Dev} & {\bf Min} & {\bf 25\%} & {\bf 50\%} & {\bf 75\%} & {\bf Max} \\
\hline
\hline
\begin{minipage}{17ex}\vspace{0.05in}Autoencoder\vspace{0.05in}\end{minipage} & {\bf 43.08} & {\bf 33.53} & {\bf 0} & 24 & {\bf 33} & {\bf 50} & {\bf 346} \\
\hline\begin{minipage}{17ex}\vspace{0.05in}Butterworth\\ Filter\vspace{0.05in}\end{minipage} & 56.05 & 52.55 & {\bf 0} & 27 & 39 & 63 & 589 \\
\hline\begin{minipage}{17ex}\vspace{0.05in}Multi-Frequency Butterworth\vspace{0.05in}\end{minipage} & 52.10 & 47.97 & {\bf 0} & 25 & 37 & 59 & 515 \\
\hline\begin{minipage}{17ex}\vspace{0.05in}biosppy\vspace{0.05in}\end{minipage} & 58.73 & 40.75 & {\bf 0} & 34 & 48 & 70 & 460 \\
\hline\begin{minipage}{17ex}\vspace{0.05in}elgendi2010\vspace{0.05in}\end{minipage} & 242.33 & 85.12 & 4 & 184 & 237 & 293 & 604 \\
\hline\begin{minipage}{17ex}\vspace{0.05in}engzeemod2012\vspace{0.05in}\end{minipage} & 75.57 & 42.57 & 2 & 45 & 68 & 96 & 404 \\
\hline\begin{minipage}{17ex}\vspace{0.05in}hamilton2002\vspace{0.05in}\end{minipage} & 206.09 & 80.13 & {\bf 0} & 152 & 197 & 252 & 574 \\
\hline\begin{minipage}{17ex}\vspace{0.05in}neurokit\vspace{0.05in}\end{minipage} & 112.37 & 69.03 & 2 & 66 & 93 & 139 & 579 \\
\hline\begin{minipage}{17ex}\vspace{0.05in}pantompkins1985\vspace{0.05in}\end{minipage} & 181.94 & 76.69 & {\bf 0} & 128 & 170 & 223 & 551 \\
\hline\begin{minipage}{17ex}\vspace{0.05in}vg\vspace{0.05in}\end{minipage} & 47.34 & 43.30 & {\bf 0} & {\bf 23} & 34 & 54 & 535 \\
\hline
\end{tabular}
\caption{Statistical summary of the ECG delineation error measured after denoising for noisy input data corrupted with LinearWander, MuscleArtifact, ShockPulses, and SineWander noise, reported for each denoising method.}
\label{table:delineated_multinoise}
\end{table}

\end{appendices}

%%%%%%%%%%%%%%%%%%%%%%%%%%%%%%%%%%%%%%%%%%%%%%%
% END
%%%%%%%%%%%%%%%%%%%%%%%%%%%%%%%%%%%%%%%%%%%%%%%
\end{document}